\documentclass{article}

\usepackage[preprint]{neurips_2026}

\usepackage[utf8]{inputenc}
\usepackage[T1]{fontenc}  
\usepackage{hyperref}  
\usepackage{url}        
\usepackage{booktabs}  
\usepackage{amsfonts}   
\usepackage{nicefrac}      
\usepackage{microtype}     
\usepackage{xcolor}        
\usepackage{graphicx}     
\usepackage{amsmath}
\usepackage{multirow}
\usepackage{longtable}
\usepackage{mathtools}
\usepackage{tabularx}
\usepackage{float}
\usepackage{listings}
\newfloat{listing}{tbp}{lol}
\floatname{listing}{Listing}
\usepackage{enumitem}
\usepackage{threeparttable}
\definecolor{ForestGreen}{RGB}{34,139,34}
\usepackage{colortbl}
\usepackage{wrapfig}
\usepackage{adjustbox}

\title{Reliable Virtual Sensing: A Multi-Domain Benchmark for Robustness Under Sensor Failures}

\author{%
Jens U. Brandt$^{1,2}$\thanks{Contact: jens\_uwe.brandt@th-koeln.de} \quad Noah C. Puetz$^{1,2}$ \quad Alexander Windmann$^{3}$ \quad \textbf{Marc Hilbert}$^{2,4}$ \\ \quad \textbf{Elena Raponi}$^2$ \quad \textbf{Thomas Bäck}$^2$ \quad \textbf{Thomas Bartz-Beielstein}$^1$\\
$^1$TH Köln \quad $^2$Leiden University \quad $^3$Helmut Schmidt University \quad $^4$Toyota Racing \\
}

\begin{document}

\maketitle

\begin{abstract}
Virtual sensing, the estimation of hard-to-measure quantities from available sensor measurements, is a critical enabler for control and monitoring in cyber-physical systems. However, when sensors fail, learning-based predictors can produce physically implausible estimates that propagate to system-level failures. We argue that real-world deployment demands robustness and introduce MuViS-C, the first multi-domain benchmark of robustness against common sensor failures in learning-based virtual sensing. Building on an existing nominal-performance benchmark and established corruption taxonomies, it covers ten sensor failure modes, from subtle drifts to catastrophic signal dropouts, at multiple severities. These are paired with complementary robustness measures capturing average error under corruption, relative degradation, and worst-case fragility. Across nine datasets from six domains, we benchmark six architectures spanning gradient-boosted trees and the major inductive biases for sequence modeling: convolution, recurrence, attention, and MLP-mixing. On the attention-based architecture, we further probe three robustification strategies. We find that (i) every model degrades substantially under corruption, becoming worse than a naïve predictor on at least one corruption setting, (ii) gradient-boosted tree ensembles achieve strong robustness, and (iii) dedicated robustification closes the gap between the attention-based architecture and the most robust models, though each strategy hurts nominal performance. The benchmark's multi-domain design proves essential, as model rankings shift across datasets, and no single domain captures the full robustness picture. MuViS-C is open-source and extensible to new datasets, failure modes, measures, and models.
\end{abstract}

\begin{figure}[htbp]
    \centering
    \includegraphics[width=0.95\textwidth]{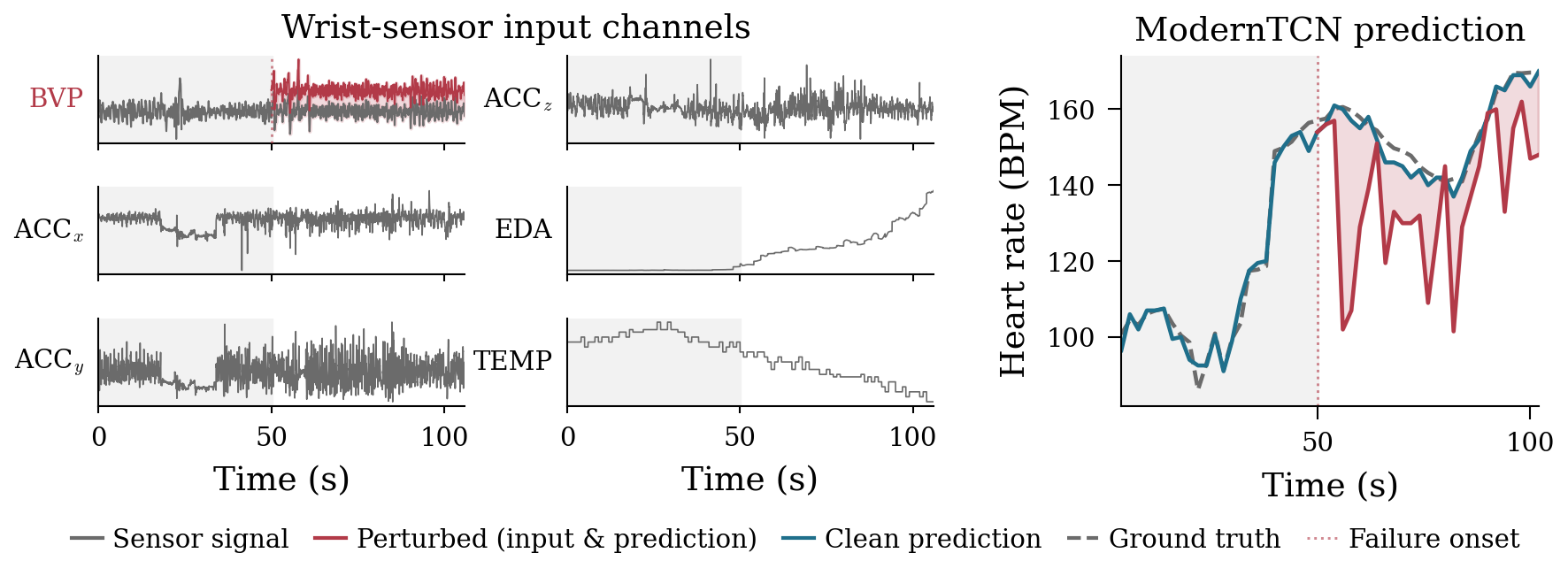}
    \vspace{-12pt}
    \caption{\textbf{Motivating Example.} An offset on the Blood Volume Pulse channel corrupts ModernTCN heart-rate estimates, causing substantial errors despite only one of six sensors being affected.}
    \label{fig:motivating_example}
\end{figure}

\section{Introduction}
\label{sec:intro}

Virtual sensing plays a central role in modern cyber-physical systems (CPS), enabling the estimation of variables from available sensors when direct measurement is technically or economically infeasible \cite{martin_virtual_2021}. With applications ranging from air quality monitoring and vehicle dynamics to industrial processes and health monitoring, virtual sensors provide a cost-effective and often safety-critical alternative to physical instrumentation \cite{ma_soft_2024, zetterberg_oskar_and_tevell_axel_creating_2023, mondal_estimating_2024, reiss_deep_2019}. These models can be physics-based, data-driven, or hybrid, depending on system knowledge and data availability. Recent efforts have standardized the evaluation of learning-based models under nominal operating conditions \cite{brandt_muvis_2026}. However, these models still depend on physical sensors as inputs and therefore inherit their most critical vulnerability: sensors fail. Thermal drift, occlusion, electrical noise, and calibration errors can corrupt measurements in real-world deployments. When such failures occur, they induce distribution shifts that learning-based models are rarely trained to handle. The result can be catastrophic errors that propagate to system-level failures \cite{brandt_faults_2025}.

Robustness to such deployment-time hazards has been identified as a core problem in ML safety \cite{hendrycks_unsolved_2022}, with standardized corruption benchmarks in computer vision \cite{hendrycks_benchmarking_2018} setting the methodological precedent. Virtual sensing, by contrast, has no comparable framework. In practice, sensor failures are rarely as simple as additive Gaussian noise or complete signal dropout. They are often subtle, structured, and temporally correlated: corrupted measurements may remain within valid physical ranges while conveying systematically misleading information. While physical sensor fault archetypes are well-catalogued \cite{balaban_modeling_2009, jesus_survey_2017} and their impact on learned models has been probed in adjacent tasks \cite{windmann_quantifying_2025, dix_measuring_2023}, virtual sensing itself lacks a multi-domain robustness benchmark. To bridge this gap, we introduce \textbf{MuViS-C}, the first standardized multi-domain benchmark of robustness against common sensor failures in learning-based virtual sensing, extending the MuViS virtual sensing benchmark \cite{brandt_muvis_2026} with a systematic robustness evaluation. Our contributions are:

\begin{itemize}
    \item \textbf{The MuViS-C benchmark.} A reproducible evaluation protocol implementing ten sensor failure modes, from subtle drifts to catastrophic signal dropouts, swept across the failure $\times$ channel $\times$ severity grid and aggregated into three robustness measures: mPC and rPC adapted from Michaelis~et~al.~\cite{michaelis_benchmarking_2020}, and $s_{\text{cross}}$, a new measure capturing worst-case fragility. The benchmark is open-source and extensible, supporting custom models, failure modes, measures, and datasets.

    \item \textbf{Multi-domain evaluation as a methodological requirement.} Across nine datasets from six domains, we show that model rankings shift substantially and no single domain captures the full robustness picture, establishing multi-domain evaluation as a requirement for assessing virtual sensor reliability.

    \item \textbf{Diagnostic study of current virtual sensors.} Benchmarking six architectures spanning gradient-boosted trees and the major inductive biases for sequence modeling (convolution, recurrence, attention, and MLP-mixing), we expose pervasive fragility under structured sensor failures and identify gradient-boosted trees as a strong implicit-robustness baseline.

    \item \textbf{Probe of robustification strategies.} On the attention-based architecture, we probe three representative robustification methods spanning input dropout, adversarial training, and purpose-built self-supervised pretraining, establishing reference points and quantifying the trade-off between nominal performance and robustness.
\end{itemize}

\section{Related work}

\paragraph{Virtual sensing.}
Methods for virtual sensing range from mechanistic white-box models, through gray-box hybrids, to fully data-driven black-box approaches that can be deployed without deep domain expertise~\cite{chen_dynamic_2020}.
While individual application domains like battery state-of-charge estimation, air-quality monitoring or vehicle dynamics have produced tailored solutions~\cite{ma_soft_2024, zetterberg_oskar_and_tevell_axel_creating_2023, mondal_estimating_2024}, cross-domain comparison has been hampered by inconsistent preprocessing, evaluation metrics, and data splits. The recently introduced MuViS benchmark addresses this gap by consolidating multimodal virtual sensing datasets from six physical domains into a unified evaluation framework~\cite{brandt_muvis_2026}. Its comparison of gradient-boosted trees and deep neural architectures shows that no single model dominates across domains, underscoring the need for architectures that generalize beyond individual applications. Crucially, however, MuViS evaluates models exclusively under nominal conditions, leaving open the question of how these architectures behave when the input sensors degrade.

\paragraph{Sensor failure modeling.}
Balaban et al.~\cite{balaban_modeling_2009} map physical failure mechanisms across eight common sensor types (thermocouples, RTDs, piezoelectric, piezoresistive, strain gages, Hall effect, magnetostrictive, and LVDTs) to five behavioral fault categories (bias, drift, scaling, noise, and hard faults), showing that these categories are tied to the sensing principle rather than the application domain. Jesus et al.~\cite{jesus_survey_2017} arrive at a similar taxonomy for wireless sensor networks, identifying six failure modes (offset, drift, crash, trimming, outliers, and noise) and surveying fusion-based mitigation strategies that exploit spatial, temporal, and value redundancy. The convergence of both taxonomies, reinforced by additional fault-type analyses \cite{ni_sensor_2009} and empirical prevalence studies in deployed sensor networks \cite{sharma_sensor_2010}, confirms that the same fault archetypes recur wherever the same transducer physics apply, from aerospace and environmental monitoring to industrial systems. Together, they supply the physically grounded fault vocabulary that any sensor-robustness benchmark must build on. 

\paragraph{Robustness benchmarks.}
A large body of work on systematic robustness evaluation under realistic input corruptions exists in the computer vision domain. ImageNet-C/P~\cite{hendrycks_benchmarking_2018} introduced a standardized suite of 15 corruption types at five severity levels and demonstrated large accuracy drops even for state-of-the-art classifiers, catalyzing a body of work on corruption-robust architectures. Analogous benchmarks exist for object detection~\cite{michaelis_benchmarking_2020}, 
segmentation~\cite{kamann_benchmarking_2020}, and 3D perception~\cite{dong_benchmarking_2023}, 
establishing corruption robustness as a first-class evaluation axis alongside clean accuracy. Recent work extends this perspective to machine learning on industrial time series data, where sensor faults can cause catastrophic degradation in learned virtual sensors~\cite{brandt_faults_2025}. Dix et al.~\cite{dix_measuring_2023} demonstrate similar sensitivity in time series classification, while Windmann et al.~\cite{windmann_quantifying_2025} report substantial accuracy losses under injected data-quality issues in forecasting pipelines. Yet virtual sensing lacks what computer vision has long established: a systematic, 
multi-domain robustness benchmark with physically grounded corruptions.

\paragraph{Robustness improvement methods.}
A broad landscape of techniques has been proposed to improve model robustness 
under distributional shift. Adversarial training augments the training set with 
worst-case perturbations to harden models against adversarial 
inputs~\cite{szegedy_intriguing_2014, goodfellow_explaining_2015}, with projected 
gradient descent as a prominent generation method~\cite{madry_towards_2018}. Certified robustness methods provide formal guarantees that predictions remain stable within a specified perturbation set~\cite{raghunathan_certified_2018, cohen_certified_2019}. Domain adaptation and domain randomization encourage invariant representations by training on diverse environments or simulated variations~\cite{ganin_domain-adversarial_2016, tobin_domain_2017}. Representation-learning approaches, from denoising autoencoders~\cite{vincent_extracting_2008} to masked autoencoders, learn features that are inherently more tolerant to input corruption. Brandt et al.~\cite{brandt_faults_2025} recently extended this idea to sensor data by designing a self-supervised masking scheme that simulates common sensor faults during pretraining, yielding representations that generalize to unseen fault types and enable robust virtual sensing in closed-loop autonomous driving. In the multi-sensor setting, Mena et al.~\cite{mena_increasing_2024} propose input sensor dropout, randomly masking entire sensor streams during training to improve robustness to missing sensors at inference time.

\section{The MuViS-C benchmark}
\label{sec:benchmark}

\begin{figure}[htbp]
    \centering
    \includegraphics[width=\textwidth]{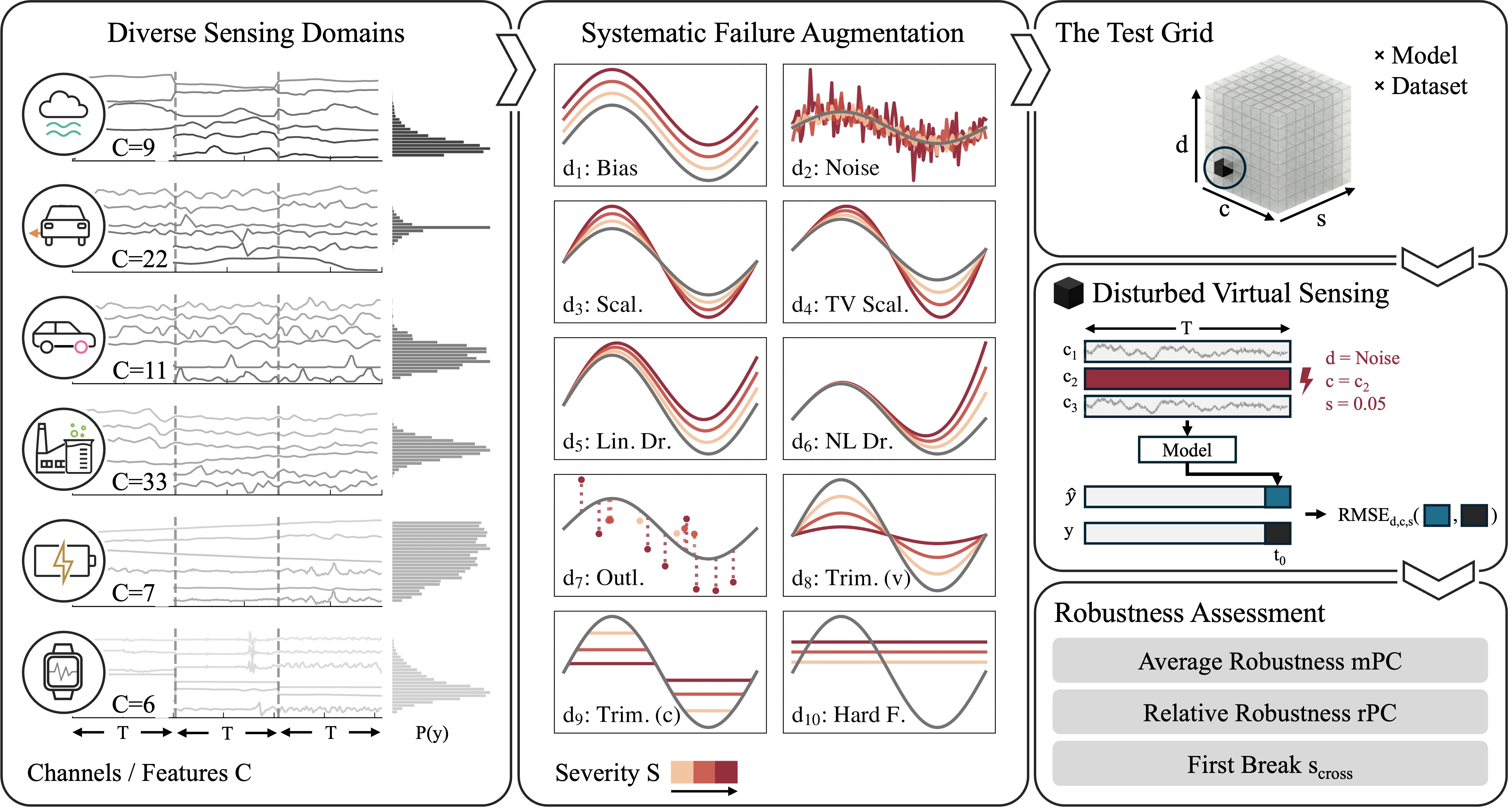}
    \caption{\textbf{MuViS-C} evaluates models across diverse sensing domains (left) by applying realistic sensor failures at increasing severity to individual channels (center). At test time, a sliding window (step size~1) feeds a $C \times T$ input tensor to the model, which produces a scalar virtual sensing estimate for~$t_0$. Each test in the failure $\times$ channel $\times$ severity grid (right top) perturbs one channel with a single failure mode at a defined severity (right center). The resulting degraded RMSEs are aggregated into complementary robustness measures (bottom right).}
    \label{fig:main_fig}
\end{figure}

We introduce MuViS-C, a multi-domain robustness benchmark for learning-based 
virtual sensing. It extends the MuViS framework~\cite{brandt_muvis_2026} 
with a systematic robustness evaluation built on physically grounded 
sensor-fault taxonomies and the corruption-benchmark methodology established 
in computer vision.
We retain its problem formulation: given an input
window $\mathbf{X}_i \in \mathbb{R}^{T \times C}$ of $C$
complementary sensor channels over $T$ time steps, a model~$f$
estimates a continuous scalar~$\hat{y}_i(t_0)$ anchored at a
reference time~$t_0$. To assess robustness, we perturb individual
channels of $\mathbf{X}_i$ with the sensor failure modes
catalogued in Section~\ref{sec:failures} and measure the resulting
change in $\hat{y}_i(t_0)$. Following Freiesleben and
Grote~\cite{freiesleben_beyond_2023}, we call a model robust if
such interventions in its input do not cause changes in
its output beyond a given tolerance. Figure~\ref{fig:main_fig} gives a high-level overview of the benchmark.

\subsection{Datasets}
\label{sec:datasets}

We evaluate on \textbf{nine} datasets spanning six application domains. Each dataset provides $N$ multivariate time-series windows $\mathbf{X}\in\mathbb{R}^{N\times T\times C}$ with scalar targets $\mathbf{y}\in\mathbb{R}^{N}$. For more details, see Appendix~\ref{app:data_details}. From environmental and health sensing, the \textbf{Beijing PM10 and PM2.5 Quality} datasets \cite{zhang_cautionary_2017} involve estimating particulate matter concentration ($10\,\mu\text{m}$ and $2.5\,\mu\text{m}$) from 9-dimensional meteorological and pollutant time series, while \textbf{PPGDalia (HR)} \cite{reiss_ppg-dalia_2019} targets heart rate estimation (BPM) from wrist-worn multimodal sensors (BVP, EDA, temperature, and acceleration). In industrial and energy systems, the \textbf{Tennessee Eastman Process (TEP)} \cite{rieth_additional_2017} represents a canonical chemical plant simulation where 33 process variables are used to infer a single chemical concentration, and the \textbf{Panasonic 18650PF (Batt.)} dataset \cite{kollmeyer_panasonic_2018} targets battery State-of-Charge (SoC) estimation from current, voltage, and temperature signals across diverse thermal conditions. For automotive and motorsport dynamics, the \textbf{REVS Motorsport} dataset \cite{kegelman_insights_2017} focuses on lateral velocity ($v_y$) estimation during high-speed racing; we evaluate three separate recording sessions with different car-track-combinations independently \textbf{(MMR/T66-13/T66-14)}. Finally, the \textbf{Vehicle Dynamics Dataset (Veh.)} \cite{mori_vehicle_2025} targets real-time tire temperature ($t_{\text{tire}}$) estimation from 11 channels capturing driver inputs and vehicle states.

\subsection{Severity-parameterized sensor failures}
\label{sec:failures}

\paragraph{Failure catalogue.}
We define a suite of ten failure modes (see Figure~\ref{fig:main_fig}, center panel and Table~\ref{tab:severity_failures}),
each acting on a single channel of the input window.
The modes are taken from a sensor fault taxonomy for virtual
sensing~\cite{brandt_faults_2025}, which defines fault functions
$D(\cdot)$ that map a nominal channel
$\mathbf{x} \in \mathbb{R}^{T}$ to a corrupted channel
$\tilde{\mathbf{x}} = D(\mathbf{x})$, with
$D(x_t)$ denoting the $t$-th entry ($t\in\{0,\dots,T-1\}$). We extend each
$D(\cdot)$ into a continuous family $D_{s,k}(\cdot)$ that maps a
severity $s \in [0,1]$ to a perturbation intensity, with $s=0$
leaving the signal unchanged (or minimally perturbed) and $s=1$ applying the mode's maximal
perturbation, bounded by a scale parameter $k$.

\paragraph{Design principle.}
All input data is z-score normalized per channel before evaluation.  On this common scale,
$k$ acts as a single, dataset-agnostic bound on the maximum
perturbation magnitude; we therefore fix $k$ across the benchmark
and omit it from the notation hereafter, writing $D_s(\cdot)$. Inspired by the classical $3\sigma$ rule used as a na\"{i}ve threshold for anomaly
detection~\cite{chandola_anomaly_2009}, we anchor $k\!=\!3$ so that
additive and stochastic perturbations at $s\!=\!1$ reach the
magnitude conventionally treated as anomalous. Each evaluation
perturbs exactly one input channel at a time while all remaining
channels retain their clean values, the standard atomic condition in
sensor-fault taxonomies~\cite{balaban_modeling_2009,
jesus_survey_2017}. This allows per-channel attribution of
degradation and enables exhaustive sweeping of the full
failure $\times$ channel $\times$ severity grid
(Section~\ref{sec:eval_protocol}).

\begin{table}[h]
  \centering
  \small
  \renewcommand{\arraystretch}{1.3}
  \setlength{\tabcolsep}{5pt}

  \caption{Severity-parameterized sensor failure modes. $p_{\max}$ is the
  outlier probability at $s=1$, and $\alpha$ the soft-clip damping factor.}
  \label{tab:severity_failures}

  \begin{tabularx}{\columnwidth}{@{} l >{\raggedright\arraybackslash}X l @{}}
    \toprule
    \textbf{Failure mode} & \textbf{Definition \(D_s(x_t)\)} & \textbf{Type} \\
    \midrule
    Bias                  & \(D_s(x_t) = x_t + s\,k\)                                                 & Additive (const.) \\
    Noise                 & \(D_s(x_t) = x_t + \varepsilon_t,\ \varepsilon_t \sim \mathcal{N}(0,(s k)^2)\) & Additive (stochastic) \\
    Scaling               & \(D_s(x_t) = x_t\bigl(1 + s(k-1)\bigr)\)                                  & Multipl.\ (const.) \\
    Time-varying scaling  & \(D_s(x_t) = x_t\bigl(1 + s(k-1)\tfrac{t}{T-1}\bigr)\)                    & Multipl.\ (time-var.) \\
    Linear drift          & \(D_s(x_t) = x_t + sk\tfrac{t}{T-1}\)                                     & Drift (linear) \\
    Non-linear drift      & \(D_s(x_t) = x_t + sk\bigl(\tfrac{t}{T-1}\bigr)^{2}\)                     & Drift (quadratic) \\

    Outliers
      & $\begin{aligned}
          &D_s(x_t) = x_t + m_t\,\delta_t, &m_t \sim \text{Ber}(sp_{\max}),\ \delta_t \sim U(-k,k)
        \end{aligned}$
      & Sporadic spikes \\

    Trimming (var.)
      & $\begin{aligned}
          &h=(1-s)k, &D_s(x_t)= \begin{cases}
            -h + \alpha(x_t + h) & x_t < -h\\
            \phantom{-}h + \alpha(x_t - h) & x_t > h\\
            x_t & |x_t|\le h
          \end{cases}
        \end{aligned}$
      & Saturation (soft) \\

    Trimming (const.)     & \(h=(1-s)k,\ D_s(x_t)=\operatorname{clip}(x_t,-h,h)\)                     & Saturation (hard) \\
    Hard fault & \(D_s(x_t) = s\,k\)                                                       & Stuck-at \\
    \bottomrule
  \end{tabularx}

  \vspace{4pt}
  \raggedright\scriptsize
  Defaults: \(k=3,\ p_{\max}=0.3,\ \alpha=0.4\).
\end{table}

\subsection{Evaluation protocol}
\label{sec:eval_protocol}

For every (dataset, model) pair, failure mode $d \in \mathcal{D}$,
and affected input channel $c \in \mathcal{C}$, we sweep severity $s$
over the grid $\mathcal{S} = \{0.05, 0.10, \dots, 1.0\}$ of
$|\mathcal{S}| = 20$ equidistant steps of size $\Delta s = 0.05$,
recording the corrupted $\mathrm{RMSE}_{d,c,s}$ at each grid point. All RMSE values are bootstrap means over 200 resamples of the test set. Two additional reference points anchor our evaluation:
$\mathrm{RMSE}_{\mathrm{clean}}$, the model's nominal RMSE on the
unperturbed test set; and $\mathrm{RMSE}_{\mathrm{naive}}$, the RMSE
of a severity-invariant baseline that always predicts the
training-set mean and depends only on the dataset. We measure robustness with three metrics: the mean Performance under
Corruption (mPC) and the relative Performance under Corruption (rPC)
from Michaelis~et~al.~\cite{michaelis_benchmarking_2020}, and the
Baseline Crossing Severity ($s_{\mathrm{cross}}$) we propose to capture
worst-case fragility. All three are defined below.

\paragraph{Mean performance under corruption (mPC).}
mPC is the average $\mathrm{RMSE}_{d,c,s}$ over the corruption grid:
\begin{equation}\label{eq:mpc}
  \mathrm{mPC}
  = \frac{1}{|\mathcal{D}|\,|\mathcal{C}|\,|\mathcal{S}|}
    \sum_{d \in \mathcal{D}}\sum_{c \in \mathcal{C}}\sum_{s \in \mathcal{S}}
    \mathrm{RMSE}_{d,c,s}\,.
\end{equation}

It instantiates the original mPC with RMSE as the performance metric, and additionally averages over the affected channel $c$. mPC is reported in the original units of the target variable and is therefore not directly comparable across datasets.

\paragraph{Relative performance under corruption (rPC).}
mPC is influenced by a model's nominal performance: a stronger model
enters the severity sweep at a lower mPC floor.
To measure \emph{degradation} independently of where a model starts, we
normalize mPC by $\mathrm{RMSE}_{\mathrm{clean}}$:
\begin{equation}\label{eq:rpc}
  \mathrm{rPC}
  = \frac{\mathrm{mPC}}{\mathrm{RMSE}_{\mathrm{clean}}}\,.
\end{equation}
$\mathrm{rPC}\!=\!1$ indicates no degradation on average, and values
above~$1$ signal performance loss.
This metric plays the same role as the Relative~mCE \cite{hendrycks_benchmarking_2018}, isolating degradation from
nominal accuracy and is the reciprocal of the robustness score of
Windmann~et~al.~\cite{windmann_quantifying_2025}
(see Appendix~\ref{app:metric_discussion}).
A complementary normalization of mPC against the na\"{i}ve mean
predictor, the normalized Performance under Corruption (nPC), is
defined and reported in Appendix~\ref{app:npc}.

\paragraph{Baseline crossing severity ($s_{\mathrm{cross}}$).}
Both mPC and rPC summarize \emph{average-case} behavior.
To capture \emph{worst-case} fragility we ask: at what severity does the
model first become no better than the na\"{i}ve mean predictor?
$s_{\mathrm{cross}}$ is the earliest severity at which \emph{any}
$\mathrm{RMSE}_{d,c,s}$ is no better than
$\mathrm{RMSE}_{\mathrm{naive}}$:
\begin{equation}\label{eq:scross}
s_{\mathrm{cross}}
  = \min_{d \in \mathcal{D},\,c \in \mathcal{C}}\;
    \min\bigl\{s \in \mathcal{S} \;\big|\;
    \mathrm{RMSE}_{d,c,s} \ge \mathrm{RMSE}_{\mathrm{naive}}\bigr\}\,.
\end{equation}
If no crossing is observed within the measured range,
$s_{\mathrm{cross}}$ is undefined and denoted by~``--'' in tables.
Unlike mPC and rPC, which average over the $(d,c,s)$ grid,
$s_{\mathrm{cross}}$ is a minimum, since averaging would mask the
single worst-case combination it is designed to detect.

\paragraph{Cross-dataset aggregation.}
We summarize mPC and rPC across datasets in two complementary ways.
Following Fleming and Wallace~\cite{fleming_how_1986}, we use the
geometric mean; rPC enters it directly as a
dimensionless ratio, whereas mPC carries incomparable units across
datasets and is first normalized per dataset by the best model
($1.0$ = best). We additionally report mean ranks following the Demšar~\cite{demsar_statistical_2006} 
protocol; statistical testing details are deferred to Appendix~\ref{app:significance}. For $s_{\mathrm{cross}}$ we report only the mean rank, since no crossing has no finite value and admits no scale-preserving
average; no crossing is encoded as a sentinel above $1$, so such
entries tie for the best rank.

\paragraph{Complementarity.}
The three metrics are designed to
be read together:
mPC measures absolute error under corruption,
rPC isolates degradation from nominal performance, and
$s_{\mathrm{cross}}$ captures worst-case fragility.
A model with low mPC but high rPC is accurate yet fragile;
low rPC but high mPC indicates robustness without strong nominal
performance;
low rPC but low $s_{\mathrm{cross}}$ reveals smooth average degradation
masking an early collapse on the weakest combination.
Reporting all three prevents any single metric from hiding the tradeoff between nominal performance and robustness.

\subsection{Standardized interface}
\label{sec:interface}

Listing~\ref{lst:muvis-rb-api} illustrates the core evaluation workflow.
Users instantiate a \texttt{Testbed} with a dataset, register one or more
models, and optionally extend the default failure suite or add custom metrics.
A single call to \texttt{run()} sweeps the full evaluation grid and returns a \texttt{Results} object for inspection and export.
A detailed description of the programming interface is provided in
Appendix~\ref{app:api}.

\begin{listing}[htbp]
\begin{lstlisting}[
  language=Python,
  basicstyle=\small\ttfamily,
  backgroundcolor=\color{gray!5},
  frame=lines,
  framesep=4pt,
  breaklines=true,
  showstringspaces=false,
  columns=fullflexible,
  keepspaces=true,
  tabsize=4,
  commentstyle=\itshape\color{gray!60!black},
  keywordstyle=\bfseries\color{blue!60!black},
  stringstyle=\color{ForestGreen},
]
import muvis_c as rob

def my_predict_fn(X: np.ndarray) -> np.ndarray: ...  # (N,T,C) -> (N,)

testbed = rob.Testbed(dataset=rob.PPGDalia)
testbed.add_model("MyModel", predict_fn=my_predict_fn)

results = testbed.run()
results.summary()
\end{lstlisting}
\caption{Evaluating a custom virtual sensing model with MuViS-C.}
\label{lst:muvis-rb-api}
\vspace{-6pt}
\end{listing}

\section{Experiments and results}
\label{sec:results}

With MuViS-C in place, we conduct a two-part empirical study. The first part 
characterizes how state-of-the-art machine learning models behave under 
sensor failure. We evaluate six baseline architectures encompassing gradient-boosted tree ensembles and representative neural inductive biases. \textbf{XGBoost}~\cite{chen_xgboost_2016} and \textbf{Catboost}~\cite{prokhorenkova_catboost_2019} are trained on flattened inputs $\mathbb{R}^{N \times (T \cdot C)}$. For sequential processing, \textbf{xLSTM-Mixer}~\cite{kraus_xlstm-mixer_2025} is a recurrent architecture that stacks scalar-memory sLSTM blocks with exponential gating to jointly mix temporal and cross-variate information by reconciling original and reversed sequence views. \textbf{TST}~\cite{zerveas_transformer-based_2021} applies a standard Transformer encoder where each token corresponds to a full time-step vector, enabling multi-head self-attention across the temporal dimension. \textbf{PatchTSMixer}~\cite{ekambaram_tsmixer_2023} segments time series into patches and applies lightweight MLP-Mixer blocks to mix information across patches, channels, and hidden features, capturing both temporal patterns and cross-variate correlations. \textbf{ModernTCN}~\cite{donghao_moderntcn_2023} modernizes temporal convolutional networks by adopting a Transformer-block-like layout that decouples temporal and feature mixing. All sequential models operate on inputs $\mathbf{X}\in\mathbb{R}^{N\times T\times C}$.

The baselines reveal where current models break; the second part of our study 
asks whether established robustification techniques can mitigate these failures, 
and at what cost. We evaluate three representative strategies: input dropout 
(\textbf{ISensD}~\cite{mena_increasing_2024}), adversarial training 
(\textbf{PGD}~\cite{madry_towards_2018}), and purpose-built self-supervised 
pretraining (\textbf{F2F}~\cite{brandt_faults_2025}). ISensD randomly masks 
entire sensor channels during training to improve robustness to missing sensors 
at inference. PGD frames robustness as a min-max optimization, generating 
adversarial perturbations via multi-step Projected Gradient Descent during 
training. F2F is a self-supervised pretraining scheme in which an encoder is 
trained to reconstruct channels corrupted by realistic sensor failure modes; a 
virtual sensor is then trained on the frozen robust embedding. All three are 
applied on top of the TST backbone: F2F's published implementation is built on 
a TST encoder, so fixing TST as the shared backbone enables a fair head-to-head 
comparison. TST is also mid-pack on nominal performance (Appendix~\ref{app:nom}) 
and among the most fragile baselines on rPC (Table~\ref{tab:rpc}), leaving clear 
headroom for robustification gains without ceiling effects. Full training, tuning, and compute details are reported in Appendix~\ref{app:implementation}.

Tables~\ref{tab:mpc} and~\ref{tab:rpc} report mPC and rPC, respectively;
Table~\ref{tab:crossing_severity} reports $s_{\mathrm{cross}}$.  Omnibus tests reach significance for mPC and rPC
(Friedman, both $p<0.001$) but not for $s_{\mathrm{cross}}$ (Friedman,
$p=0.187$). A na\"ive-normalized variant (nPC) is reported in
Appendix~\ref{app:npc} for interpretability.  Full statistical details,
including all $p$-values, critical distances, and post-hoc cluster
memberships, are given in Appendix~\ref{app:significance}.

\begin{table*}[h]
\caption{mPC per model and dataset. Bold = best, \underline{underline} = second-best per column. GM is the geometric mean of best-normalized values (1.0 = best); arrows show GM change vs. TST.}
\label{tab:mpc}
\begin{center}
\setlength{\tabcolsep}{1.5pt}
\adjustbox{max width=\textwidth}{%
\begin{tabular}{l c c c c c c c c c c l}
\toprule
\textbf{Model} & \textbf{PM10} & \textbf{PM2.5} & \textbf{Batt.} & \textbf{HR} & \textbf{TEP} & \textbf{Veh.} & \textbf{MMR} & \textbf{T66-13} & \textbf{T66-14} & \textbf{Rank} & \textbf{GM} \\
\midrule
XGBoost & \textbf{97.12} & \underline{66.32} & 0.05 & 12.38 & \textbf{0.05} & \textbf{4.44} & \underline{0.17} & \underline{0.12} & \textbf{0.11} & \textbf{2.3} & \underline{1.202} \\
CatBoost & 98.81 & 67.51 & \underline{0.04} & 11.46 & \textbf{0.05} & 5.00 & \textbf{0.16} & \textbf{0.11} & \textbf{0.11} & 2.6 & \textbf{1.162} \\
M-TCN & 98.73 & 69.00 & 0.08 & \textbf{5.27} & \underline{0.06} & 11.39 & 0.22 & 0.14 & 0.15 & 5.6 & 1.403 \\
P-TSMixer & 107.16 & 74.35 & 0.09 & 10.25 & \textbf{0.05} & 7.92 & 0.34 & 0.21 & 0.23 & 7.2 & 1.705 \\
xLSTM-M & \underline{98.33} & 67.26 & 0.05 & \underline{9.65} & \textbf{0.05} & \underline{4.47} & 0.18 & \underline{0.12} & 0.13 & 3.0 & 1.209 \\
TST & 106.23 & 73.51 & 0.06 & 12.52 & \textbf{0.05} & 8.17 & 0.21 & 0.14 & 0.16 & 6.0 & 1.476 \\
\midrule
TST-PGD & 101.39 & \textbf{65.98} & 0.07 & 15.36 & \textbf{0.05} & 4.57 & 0.25 & 0.15 & 0.16 & 5.6 & 1.430 {\scriptsize\textcolor{ForestGreen}{$\downarrow$-0.046}} \\
TST-ISensD & 108.56 & 74.41 & 0.11 & 15.33 & \textbf{0.05} & 5.66 & 0.24 & 0.17 & 0.19 & 7.7 & 1.618 {\scriptsize\textcolor{red}{$\uparrow$+0.142}} \\
TST-F2F & 118.44 & 68.75 & \textbf{0.03} & 23.53 & 0.07 & 4.96 & 0.18 & \underline{0.12} & \underline{0.12} & 5.1 & 1.318 {\scriptsize\textcolor{ForestGreen}{$\downarrow$-0.158}} \\
\bottomrule
\end{tabular}
}
\end{center}
\end{table*}

\begin{table*}[h]
\caption{rPC per model and dataset. Bold = best, \underline{underline} = second-best per column. GM is the geometric mean across datasets; arrows indicate GM change relative to TST.}
\label{tab:rpc}
\begin{center}
\setlength{\tabcolsep}{1.5pt}
\adjustbox{max width=\textwidth}{%
\begin{tabular}{l c c c c c c c c c c l}
\toprule
\textbf{Model} & \textbf{PM10} & \textbf{PM2.5} & \textbf{Batt.} & \textbf{HR} & \textbf{TEP} & \textbf{Veh.} & \textbf{MMR} & \textbf{T66-13} & \textbf{T66-14} & \textbf{Rank} & \textbf{GM} \\
\midrule
XGBoost & 1.060 & 1.087 & \textbf{2.159} & 1.301 & \underline{1.007} & 1.187 & 1.395 & 1.211 & 1.425 & 3.4 & 1.281 \\
CatBoost & 1.073 & 1.100 & 3.669 & 1.270 & \underline{1.007} & 1.302 & 1.517 & 1.263 & 1.353 & 4.1 & 1.385 \\
M-TCN & 1.086 & 1.149 & 9.200 & 2.187 & 1.059 & 4.837 & 1.853 & 1.854 & 1.896 & 8.1 & 2.112 \\
P-TSMixer & \underline{1.043} & 1.108 & 2.613 & \textbf{1.099} & \textbf{1.006} & 2.043 & 1.247 & 1.509 & 1.559 & 3.8 & 1.395 \\
xLSTM-M & \textbf{1.042} & \textbf{1.061} & 5.550 & 1.309 & \underline{1.007} & \underline{1.178} & 1.777 & 1.645 & 1.634 & 4.7 & 1.529 \\
TST & 1.091 & 1.155 & 3.761 & 1.817 & 1.010 & 2.157 & 1.897 & 1.858 & 1.948 & 7.9 & 1.716 \\
\midrule
TST-PGD & 1.070 & \underline{1.066} & 2.833 & \underline{1.175} & \textbf{1.006} & 1.324 & \textbf{1.182} & \textbf{1.108} & \underline{1.147} & \textbf{2.7} & \textbf{1.253} {\scriptsize\textcolor{ForestGreen}{$\downarrow$-0.463}} \\
TST-ISensD & 1.134 & 1.105 & 4.129 & 1.272 & 1.011 & 1.217 & 1.525 & 1.608 & 1.736 & 6.2 & 1.482 {\scriptsize\textcolor{ForestGreen}{$\downarrow$-0.234}} \\
TST-F2F & 1.095 & 1.102 & \underline{2.208} & 1.521 & 1.231 & \textbf{1.159} & \underline{1.202} & \underline{1.126} & \textbf{1.113} & 4.1 & \underline{1.272} {\scriptsize\textcolor{ForestGreen}{$\downarrow$-0.444}} \\
\bottomrule
\end{tabular}
}
\end{center}
\end{table*}

\subsection{Gradient-boosted trees as a strong robustness baseline}

Gradient-boosted trees hold up well under corruption without any dedicated defense. On mPC (Table~\ref{tab:mpc}), XGBoost achieves the best mean rank (2.3) with CatBoost a close second (2.6), both significantly outranking P-TSMixer and TST-ISensD. On rPC, XGBoost again leads all non-robustified models, trailing only the adversarially robustified TST-PGD (Table~\ref{tab:rpc}). XGBoost is in fact the only model in the benchmark that holds an exclusive top-cluster position on both statistics (Appendix~\ref{app:significance}). No neural architecture reaches the joint top position, even with dedicated robustification. This suggests that the inductive biases of gradient-boosted trees confer an implicit corruption resilience that neural architectures in our benchmark do not replicate, setting a strong baseline to beat.

\subsection{Trade-off between nominal performance and robustness}
\label{sec:tradeoff}

\begin{wrapfigure}{r}{0.45\textwidth}
  \vspace{-12pt}
  \centering
  \includegraphics[width=0.44\textwidth]{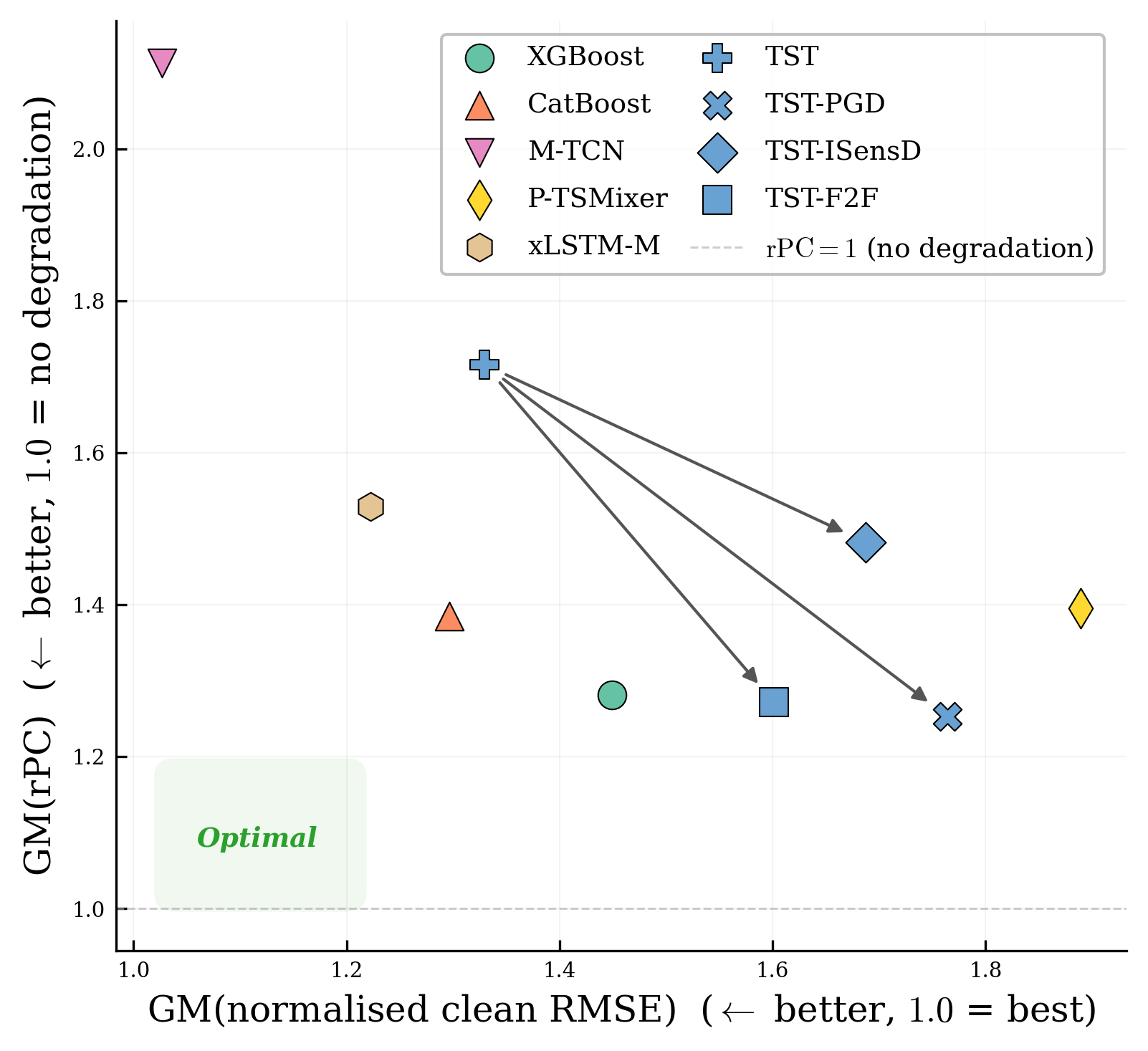}
  \caption{Nominal performance vs.\ robustness. GM(rPC) aggregated across datasets ($y$) vs.\ GM of normalized clean RMSE ($x$, normalized following mPC aggregation). Arrows indicate the effect of robustification.}
  \label{fig:robustness-vs-nominal-wrap}
  \vspace{-10pt}
\end{wrapfigure}

Figure~\ref{fig:robustness-vs-nominal-wrap} plots each model's clean-data error against its rPC, both as geometric means across datasets. The clean-data axis is normalized per dataset against the best model, so 1.0 marks the best achievable nominal performance and the optimal corner sits at the bottom-left where a model is both accurate and degrades gracefully. M-TCN, the strongest model nominally (Appendix~\ref{app:nom}), is the least
robust, with an rPC (Table~\ref{tab:rpc}) of 2.112 and the worst mean rank (8.1), significantly outranked by TST-PGD, XGBoost, and P-TSMixer. P-TSMixer shows the inverse pattern: among non-robustified baselines, it ranks among the weakest on clean-data error yet sits in the top non-significant cluster on rPC (Figure~\ref{fig:cd_rpc}), degrading gracefully from a poor nominal baseline. Among non-robustified models, xLSTM-M comes closest to the tree ensembles on 
both axes. It is also the only one to stay below the naïve-predictor threshold on PM10 and HR; all other non-robustified models cross on every dataset (Table~\ref{tab:crossing_severity}).

A genuine trade-off emerges in robustification training. The arrows in Figure~\ref{fig:robustness-vs-nominal-wrap} trace each robustified TST variant's position relative to vanilla TST, and all three point toward lower rPC but higher clean-data error: every robustification method we evaluated trades nominal performance for robustness. Statistically, vanilla TST suffers a significant rPC deficit relative to TST-PGD, XGBoost, and P-TSMixer, and all three robustification strategies close it. Every robustified variant joins the leading non-significant rPC cluster, and TST-PGD significantly outranks vanilla TST itself, though none establishes a significant advantage over the strongest non-robustified baselines (Figure~\ref{fig:cd_rpc}).

\begin{table*}[h]
\caption{Baseline crossing severity $s_{\mathrm{cross}}$. ``--'' denotes never crosses within the measured severity range (best possible). Bold = best, \underline{underline} = second-best per column.}

\label{tab:crossing_severity}
\begin{center}
\setlength{\tabcolsep}{3pt}
\adjustbox{max width=\textwidth}{%
\begin{tabular}{l c c c c c c c c c c}
\toprule
\textbf{Model} & \textbf{PM10} & \textbf{PM2.5} & \textbf{Batt.} & \textbf{HR} & \textbf{TEP} & \textbf{Veh.} & \textbf{MMR} & \textbf{T66-13} & \textbf{T66-14} & \textbf{Rank} \\
\midrule
XGBoost & 0.500 & 0.400 & 0.100 & 0.500 & 0.250 & 0.050 & 0.150 & 0.100 & 0.150 & 5.6 \\
CatBoost & 0.550 & 0.350 & 0.100 & \underline{0.950} & 0.300 & \underline{0.100} & 0.100 & 0.150 & 0.100 & 4.9 \\
M-TCN & 0.450 & \underline{0.600} & 0.450 & 0.450 & 0.050 & 0.050 & 0.050 & 0.100 & 0.050 & 5.9 \\
P-TSMixer & \underline{0.650} & 0.450 & 0.400 & 0.050 & 0.300 & 0.050 & 0.050 & 0.050 & 0.050 & 6.2 \\
xLSTM-M & \textbf{--} & 0.550 & \underline{0.600} & \textbf{--} & 0.300 & \underline{0.100} & 0.150 & 0.150 & 0.050 & \textbf{3.5} \\
TST & 0.250 & 0.200 & 0.450 & 0.500 & \underline{0.500} & 0.050 & 0.050 & 0.050 & 0.050 & 6.2 \\
\midrule
TST-PGD & 0.050 & 0.400 & 0.400 & 0.450 & \textbf{0.800} & 0.050 & \underline{0.400} & \underline{0.600} & \underline{0.450} & 4.4 \\
TST-ISensD & 0.300 & 0.350 & 0.200 & 0.300 & 0.350 & \textbf{0.200} & \underline{0.400} & 0.400 & 0.350 & 4.4 \\
TST-F2F & 0.200 & \textbf{0.950} & \textbf{--} & 0.050 & 0.050 & \underline{0.100} & \textbf{--} & \textbf{--} & \textbf{--} & 3.7 \\
\bottomrule
\end{tabular}
}
\end{center}
\end{table*}

\subsection{Benchmark diversity across datasets}
\label{sec:diversity}
 
The nine datasets span a wide difficulty spectrum, validating the
multi-domain design.  On rPC (Table~\ref{tab:rpc}), TEP is by far the easiest: six of nine
models stay within $1\%$ of clean performance, while Battery is
universally challenging, with every model at least doubling its clean
error. Crucially, model rankings shift across datasets: M-TCN achieves the
best mPC (Table~\ref{tab:mpc}) on heart rate but the worst on Vehicle Dynamics; TST-F2F
dominates Batt. and REVS on $s_{\mathrm{cross}}$
yet collapses at the lowest severity on HR and TEP (Table~\ref{tab:crossing_severity}).  No single
dataset captures the full picture, and a benchmark restricted to any
one domain would miss failure patterns that are prominent in others.

\subsection{Additional analyses}
\label{sec:additional}
 
\paragraph{Per-failure-mode heterogeneity.}

Disaggregating rPC by failure mode (Appendix~\ref{app:per_fm}, Table~\ref{tab:rpc_fm_sorted}) reveals that corruption difficulty varies substantially: the geometric-mean rPC across models ranges from 1.073 for varying trimming to 1.922 for hard faults, with hard faults producing roughly an order of magnitude more degradation than the easiest mode. All models agree on the easiest mode (varying trimming) and all but TST-ISensD and TST-F2F agree on the hardest (hard fault); for those two, noise is the worst case. Although models broadly agree at the extremes of difficulty, the best model varies mode by mode, and the per-mode winner is always either XGBoost or one of the robustified TST variants: TST-F2F leads on bias, hard faults, and linear and nonlinear drift; TST-ISensD leads on both trimming variants; TST-PGD wins on both scaling variants; and XGBoost leads on outliers and noise. No model ranks best on more than four of the ten failure modes, reinforcing the value of multi-failure evaluation.
 
\paragraph{Robustness under detectable sensor faults.}
When a reliable failure detector is available and the operator or a system can
replace a faulty channel with its training-set mean (hard fault at
$s\!=\!0$), TST-ISensD achieves a geometric-mean $\mathrm{rPC}$ of
$1.034$ and a perfect mean rank of $1.0$ across all nine datasets,
indicating virtually no degradation and significantly outranking
CatBoost, xLSTM-M, TST, and M-TCN
(Appendix~\ref{app:isensd_discussion}).  This result follows directly
from ISensD's training procedure, which randomly zeroes out input
channels and thus explicitly prepares the model for this intervention.
The finding suggests that pairing a simple input-dropout augmentation
with a failure detection mechanism is a practical strategy
for maintaining prediction quality under sensor faults.

\section{Limitations and future work}
\label{sec:limitations}

First, our ten-mode failure catalogue, grounded in established
sensor-fault taxonomies~\cite{balaban_modeling_2009, jesus_survey_2017},
covers the dominant archetypes recurring across transducer physics
but cannot be exhaustive; MuViS-C's injection framework is designed
to absorb new modes as they are characterized, and we plan to
broaden the catalogue as the community surfaces additional failure
patterns from field deployments. Second, we evaluate three representative robustification methods on
a shared TST backbone to enable a fair head-to-head comparison with
F2F~\cite{brandt_faults_2025}; broader families of defenses,
including techniques from the vision-robustness literature, and the
question of whether our findings transfer across backbones remain
open, and we plan to address both in future work using MuViS-C as
the shared testbed. Third, our evaluation perturbs one sensor at a time; extending MuViS-C with a multi-sensor
axis covering correlated-failure schedules and channel-coverage
sweeps is a natural next step that we plan to support in future
versions of the benchmark. Fourth, we fix the perturbation scale at $k=3$ throughout the study, motivated 
by the $3\sigma$ convention on z-scored data; the sensitivity of model rankings 
to alternative scales remains an open question and a clear next experiment.

\section{Conclusion}
MuViS-C provides a testbed for evaluating learning-based virtual sensor reliability under realistic sensor failures. Its multi-domain design proves essential: datasets span a wide difficulty spectrum, model rankings shift across domains, and no single dataset captures the full picture. We find that (i)~every model evaluated degrades substantially under corruption, becoming worse than a na\"ive predictor on at least one failure-mode/channel combination; (ii)~gradient-boosted tree ensembles achieve strong robustness without any explicit defense, setting a challenging baseline for neural methods; (iii)~dedicated robustification training can elevate a fragile neural baseline into the leading robustness cluster, though every method incurs a clean-data error penalty; and (iv)~when reliable failure detection and input intervention are available, a simple input-dropout augmentation (ISensD) reduces average degradation to near zero. We release MuViS-C as an open-source, extensible platform and welcome contributions of new datasets, failure modes, measures, and models.

\newpage

\begin{ack}
Funded by the European HORIZON-KDT-JU-2023-2-RIA, project ShapeFuture, grant No 101139996 and by the German Federal Ministry BMFTR within the funding measure \emph{Forschung an Fachhochschulen -- KI-Nachwuchs@FH 2-2021} under the project \emph{TH Köln -- Künstliche Intelligenz plus (THK-KIplus)}, funding code 13FH007KI2. The authors are responsible for the content of this publication.
\end{ack}

\bibliography{MuViS-Rob}
\bibliographystyle{unsrt}

\newpage
\appendix

\section{Data and preprocessing}
\label{app:data_details}

The dataset selection, train/test splits, and preprocessing pipeline
(cleaning, resampling, channel selection, and sliding-window construction)
are taken unchanged from the MuViS benchmark~\cite{brandt_muvis_2026}; we refer the
reader to that paper for the full design rationale and summarize the
resulting dataset statistics in Table~\ref{tab:datasets}.

\begin{table}[htbp]
  \centering
  \small
  \renewcommand{\arraystretch}{1.0}
  \setlength{\tabcolsep}{10pt} 
  
  \caption{Benchmark datasets. $C$: features, $T$: sequence length, $N$: number of samples.}
  \label{tab:datasets}
  
  \begin{tabular}{llcccc}
    \toprule
    \textbf{Dataset} & \textbf{Domain} & $C$ & $T$ & $N_{\text{train}}$ & $N_{\text{test}}$ \\
    \midrule
    BeijingPM10Quality      & Air quality     &  9 &  24 & 11,918  & 5,048   \\
    BeijingPM25Quality      & Air quality     &  9 &  24 & 11,918  & 5,048   \\
    Panasonic18650PFData    & Battery SoC     &  7 & 120 & 199,827 & 158,126 \\
    PPGDalia                & Wearable / Bio  &  6 & 512 & 51,757  & 12,940  \\
    REVS/2013 Monterey      & Motorsport      & 22 &  20 & 120,777 & 21,357  \\
    REVS/2013 Targa 66      & Motorsport      & 22 &  20 & 33,520  & 11,109  \\
    REVS/2014 Targa 66      & Motorsport      & 22 &  20 & 46,739  & 9,036   \\
    TennesseeEastman        & Chemical        & 33 &  20 & 240,500 & 470,500 \\
    VehicleDynamics         & Automotive      & 11 &  50 & 1,384   & 280     \\
    \bottomrule
  \end{tabular}
\end{table}

\section{Implementation details}
\label{app:implementation}

Each of the nine model classes was
trained on each of the nine benchmark datasets, yielding 81 (model~$\times$
~dataset) checkpoints used throughout the evaluation. All hyperparameters
are reported in the accompanying repository.

\subsection{Baselines}

All neural networks consume tensors in the $(N, T, C)$ layout and emit a
single scalar per window. Apart from TST, which targets representation
learning for regression and classification, the architectures were
originally proposed for multivariate forecasting and required minor
adjustments for the scalar virtual-sensing setting.

\paragraph{TST.} Used as-is with the built-in linear regression head
producing a single output.

\paragraph{ModernTCN.} Instantiated in regression mode with a single output
unit; RevIN is disabled. Inputs are transposed to $(N, C, T)$ to match the original
implementation.

\paragraph{xLSTM-Mixer.} The forecast horizon is set to zero, and predictions
are obtained through a regression pathway (mirroring the model's
classification adaptation in the released code) that flattens the
post-mixer tokens through a single linear head.

\paragraph{PatchTSMixer.} The Hugging Face implementation of
\texttt{PatchTSMixerForRegression} is used with \texttt{num\_targets=1},
\texttt{mode="common\_channel"}, and \texttt{head\_aggregation="use\_last"}.

\subsection{Optimization}

All neural networks are trained for 100 epochs with Adam, batch size 256,
and MSE loss. Inputs are z-score normalized per feature using statistics
estimated on the 90\% training portion of an additional 90/10 train/val
split. Checkpoints are selected by best validation loss, and test metrics
are reported on the predefined temporal test split. All runs use seed~42.

\subsection{Robustification methods}

\paragraph{F2F (self-supervised pretraining + fine-tuning).} The encoder is
pretrained for 600 epochs with a multi-task masked-reconstruction objective.
For each batch, samples are split into three disjoint subsets, and the
positions selected by a geometric binary mask are corrupted by
(i)~zeroing (mean replacement on z-scored data),
(ii)~additive uniform bias, or
(iii)~additive Gaussian noise; the three task losses are summed with equal
weights. Pretraining uses AdamW with separated decay/no-decay parameter
groups, a cosine schedule with linear warmup, gradient clipping (after
warmup), and zero weight decay. A two-layer MLP projection head maps the
flattened encoder output $(T \cdot d_{\text{model}})$ back to the
reconstruction target. After pretraining, the projection head is discarded,
the regression head is reinitialized, the backbone is frozen, and only the
head is fine-tuned for 100 epochs with the same supervised setup as the
TST baseline.

\paragraph{PGD (adversarial training).} Each batch is split into a clean
portion and an adversarial portion, where the
adversarial fraction $\rho$ is tuned per dataset. Adversaries are generated
with PGD under an $\ell_\infty$ budget for a fixed number of iterations,
initialized from a random point inside the $\epsilon$-ball. All other optimization settings match the TST baseline.

\paragraph{ISensD (input sensor dropout).} For each training sample, a
non-empty subset of input channels is sampled uniformly from the $2^C - 1$
possibilities and the remaining channels are zeroed before the forward
pass. The model is therefore trained over the full distribution of
sensor-availability patterns rather than a fixed configuration.
Optimization is otherwise identical to the TST baseline.

\subsection{Compute resources}

All experiments were implemented in Python~3.13.7 and executed on a single
NVIDIA H100 (80~GB) GPU with CUDA~13.2. To accelerate evaluation, compatible models were compiled with \texttt{torch.compile} and inference was run under \texttt{torch.autocast} (\texttt{bfloat16}). Full reproduction of the paper, covering training for all 81 (model~$\times$~dataset) configurations followed by the complete
severity sweep, takes approximately 155 hours of GPU time on a single
device, though training and evaluation are embarrassingly parallel
across (model, dataset) configurations and can be spread across
multiple GPUs for substantial wall-clock speedups. Hyperparameter
tuning and preliminary experiments leading up to the final results
required considerably more compute.

\subsection{Hyperparameter selection}
\label{app:hp}
 
\subsubsection{Baseline architectures}
Hyperparameters for XGBoost and CatBoost were adopted directly from the
MuViS benchmark~\cite{brandt_muvis_2026}.
For the remaining baselines (TST, ModernTCN, PatchTSMixer, xLSTM-Mixer),
we followed a comparable protocol: Optuna~\cite{akiba_optuna_2019} with
multivariate TPE sampling, 100~trials per architecture and dataset,
minimizing validation RMSE on a 10\,\% subsample of the training data.
Where available, the first trial was initialized with
author-recommended defaults.
Table~\ref{tab:hp_baselines} lists the search spaces.

\begin{table}[h]
\centering
\small
\caption{Search spaces for baseline architectures. Brackets denote continuous ranges; braces denote categorical choices.}
\label{tab:hp_baselines}
\begin{tabular}{llll}
\toprule
\textbf{Model} & \textbf{Parameter} & \textbf{Range / Choices} \\
\midrule
\multirow{6}{*}{TST}
  & $d_\text{model}$       & \{64, 128, 256\} \\
  & $n_\text{heads}$       & \{4, 8, 16\} \\
  & num\_layers            & [1, 4] \\
  & dim\_feedforward       & \{128, 256, 512\} \\
  & dropout                & [0.0, 0.2] \\
  & learning rate          & [$10^{-5}$, $10^{-3}$] (log) \\
\midrule
\multirow{7}{*}{ModernTCN}
  & $d_\text{model}$       & \{32, 64, 128\} \\
  & num\_blocks            & [1, 4] \\
  & large\_kernel          & \{7, 13, 25, 31, 51, 71\} \\
  & small\_kernel$^\dagger$ & \{3, 5, 7, 13, 25\} \\
  & ffn\_ratio             & \{1, 2, 4\} \\
  & dropout / head\_dropout & [0.0, 0.3] \\
  & learning rate          & [$10^{-5}$, $10^{-2}$] (log) \\
\midrule
\multirow{6}{*}{PatchTSMixer}
  & patch\_length$^\ddagger$ & \{2, 4, 5, 8, 10, 16, 20, 32, 64\} \\
  & $d_\text{model}$       & \{32, 64, 128\} \\
  & num\_layers            & [2, 8] \\
  & expansion\_factor      & \{2, 4\} \\
  & dropout / head\_dropout & [0.0, 0.3] \\
  & learning rate          & [$10^{-5}$, $10^{-2}$] (log) \\
\midrule
\multirow{7}{*}{xLSTMMixer}
  & embedding\_dim         & \{128, 256, 512\} \\
  & num\_heads             & \{4, 8, 16\} \\
  & num\_blocks            & [1, 4] \\
  & num\_mem\_tokens       & \{0, 2, 4, 8\} \\
  & conv1d\_kernel\_size$^\S$ & \{0, 2, 4, 8, 16, 32\} \\
  & dropout                & [0.0, 0.3] \\
  & learning rate          & [$10^{-5}$, $10^{-2}$] (log) \\
\bottomrule
\end{tabular}
\begin{tablenotes}
\small
\item[$\dagger$] Constrained to $\leq$ large\_kernel.
\item[$\ddagger$] Constrained to values that evenly divide the sequence length and yield $\geq 4$ patches.
\item[$\S$] Constrained to $\leq$ sequence length.
\end{tablenotes}
\end{table}

\subsubsection{Robustness methods}
\label{app:hp_robustness}
 
All three robustness variants use the TST architecture with the
per-dataset hyperparameters found above; only the
method-specific parameters differ.
 
\paragraph{ISensD.}
Input-Sensor Dropout is parameter-free: during training, each input
channel is independently zeroed out with a fixed probability.
No additional tuning is required beyond the base TST configuration.
 
\paragraph{F2F.}
We adopt the optimizer settings and masking parameters from the
original publication~\cite{brandt_faults_2025}, adjusting only the average mask
length to match the sequence length of each dataset.
No further search is performed.
 
\paragraph{PGD.}
Adversarial training with Projected Gradient Descent introduces four
parameters: the perturbation budget~$\epsilon$, the PGD step
size~$\alpha$, the number of inner steps~$K$, and the fraction of
samples in each batch that are adversarially perturbed,
$r_{\mathrm{adv}}$. Because these parameters control the strength of the adversarial
perturbation rather than the model architecture, tuning them on clean
validation loss alone risks selecting configurations that merely
regularize the model without genuinely improving robustness.
We therefore optimize a \emph{robustness-aware objective} on the
validation set.
 
\paragraph{Robustness-aware tuning objective.}
The baseline architectures are tuned on clean validation RMSE,
since the goal of the first part of our study is to characterize the
robustness of state-of-the-art models tuned for nominal performance.
For the robustification methods, an analogous clean-data objective
would be self-defeating: the tuner would simply minimize the
robustness hyperparameters that enlarge perturbations during
training (e.g., the PGD budget $\epsilon$), recovering near-vanilla
training. We therefore tune these methods against a robustness-aware
proxy that jointly rewards low clean error and stability under a
small set of corruptions. For each trial we train on a 10\,\% subsample of the training data and
evaluate a lightweight robustness proxy~$R$ on the validation set.
$R$ is the geometric mean of the per-corruption relative performance ratios
$u_{\mathrm{rel}} = \mathrm{RMSE}_{\mathrm{clean}} \,/\,
\mathrm{RMSE}_{\mathrm{corr}}$ computed over a minimal failure subset:
bias and noise at severities $\{0.33,\allowbreak 0.66,\allowbreak
1.0\}$ plus the hard-fault at severity~$0$ (stuck-at-mean), applied to
a single randomly selected feature (fixed per study).
The tuning objective is
$\mathrm{RMSE}_{\mathrm{clean}} \,/\, R$,
which jointly rewards low clean error and high robustness.
We run 50~Optuna trials with multivariate TPE sampling; the first
trial is seeded with author-recommended defaults.
Table~\ref{tab:hp_pgd} lists the search space.
 
\begin{table}[h]
\centering
\small
\caption{Search space for PGD adversarial training parameters.}
\label{tab:hp_pgd}
\begin{tabular}{ll}
\toprule
\textbf{Parameter} & \textbf{Choices} \\
\midrule
$\epsilon$ (perturbation budget) & \{0.05, 0.1, 0.2, 0.3\} \\
$\alpha$ (step size)             & \{0.01, 0.05, 0.1\} \\
$K$ (inner steps)                & \{7, 20\} \\
$r_{\mathrm{adv}}$ (adversarial batch ratio) & \{0.25, 0.5\} \\
\bottomrule
\end{tabular}
\end{table}
 
\paragraph{Design rationale.}
The tuning failure subset, bias, noise, and hard-fault at
severity~$0$, deliberately mirrors the corruption types used by F2F
during training, ensuring a comparable level of corruption exposure
across methods.
Crucially, this subset is a strict subset of the full benchmark
evaluation grid: it covers only three of the failure modes and a small
number of severities, and perturbs a single feature rather than all
channels.
This limits the overlap between the tuning signal and the test-time
evaluation, reducing the risk of overfitting to the benchmark while
still providing a meaningful robustness gradient for the optimizer.
We recommend this minimal-subset protocol for future robustness tuning
in order to balance informativeness with test-set integrity.

\section{Examples}
\label{app:examples}

\begin{figure}[h]
    \centering
    \includegraphics[width=\linewidth]{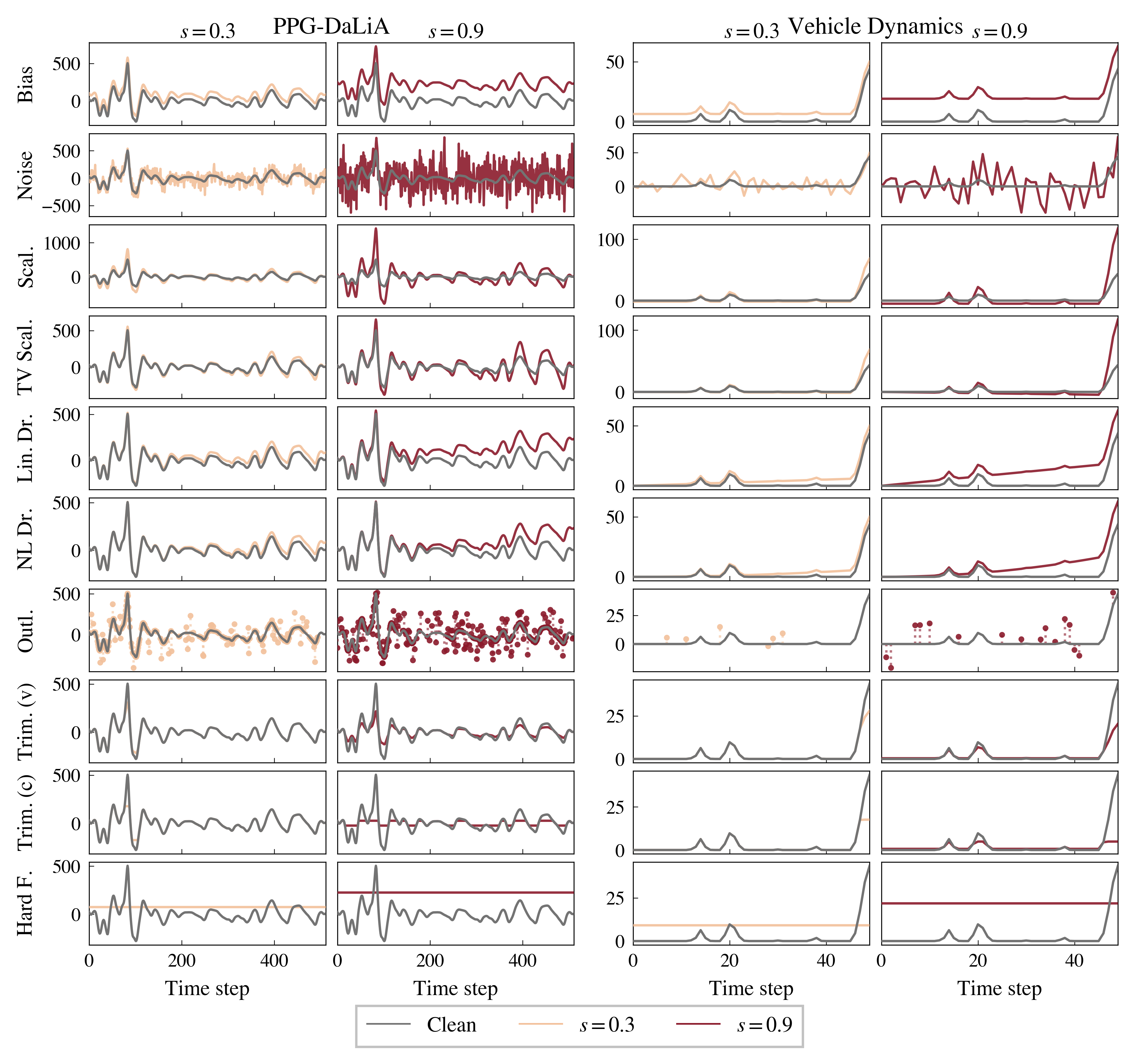}
     \caption{Illustration of the ten sensor failure modes applied to a 
    representative input window from PPG-DaLiA (left) and Vehicle 
    Dynamics (right). Each failure is shown at two severity levels 
    alongside the clean signal, highlighting the 
    qualitative differences between additive, multiplicative, 
    drift-based, saturation-based, and stuck-at corruptions.}
    \label{fig:failure_modes_examples}
    \vspace{-10pt}
\end{figure}

To give the reader an intuitive sense of how the severity-parameterized 
failure modes of Section~\ref{sec:failures} act on real data, 
Figure~\ref{fig:failure_modes_examples} visualizes each of the ten 
failure modes applied at two severity levels to a single input channel 
from the PPG-DaLiA and Vehicle Dynamics datasets.

\section{Nominal performance}
\label{app:nom}

Table~\ref{tab:nominal_performance_ci} reports clean-data RMSE for every
(model, dataset) pair. Because per-dataset RMSE spans four orders of
magnitude, we aggregate with the best-normalized geometric mean
($\mathrm{GM}$). ModernTCN leads on six of nine datasets ($\mathrm{GM}=1.027$, mean rank $2.2$),
and all three robustness-aware TST variants pay a clean-data penalty relative to the vanilla backbone
($+0.27$ to $+0.44$ in $\mathrm{GM}$).

\begin{table*}[h]
\caption{Nominal RMSE with 95\% percentile bootstrap confidence intervals (test-set resampling, $n_\text{boot}=200$). Each cell stacks the upper CI (top, grey), the bootstrap mean (centre, bold if best), and the lower CI (bottom, grey).}
\label{tab:nominal_performance_ci}
\begin{center}
\setlength{\tabcolsep}{2pt}
\renewcommand{\arraystretch}{1.10}
\adjustbox{max width=\textwidth}{%
\begin{tabular}{l c c c c c c c c c c l}
\toprule
\textbf{Model} & \textbf{PM10} & \textbf{PM2.5} & \textbf{Batt.} & \textbf{HR} & \textbf{TEP} & \textbf{Veh.} & \textbf{MMR} & \textbf{T66-13} & \textbf{T66-14} & \textbf{Rank} & \textbf{GM} \\
\midrule
XGBoost & \begin{tabular}[c]{@{}c@{}}{\footnotesize\textcolor{gray}{98.097}} \\ 91.610 \\ {\footnotesize\textcolor{gray}{85.757}}\end{tabular} & \begin{tabular}[c]{@{}c@{}}{\footnotesize\textcolor{gray}{67.091}} \\ 61.040 \\ {\footnotesize\textcolor{gray}{56.036}}\end{tabular} & \begin{tabular}[c]{@{}c@{}}{\footnotesize\textcolor{gray}{0.025}} \\ 0.025 \\ {\footnotesize\textcolor{gray}{0.025}}\end{tabular} & \begin{tabular}[c]{@{}c@{}}{\footnotesize\textcolor{gray}{9.727}} \\ 9.511 \\ {\footnotesize\textcolor{gray}{9.300}}\end{tabular} & \begin{tabular}[c]{@{}c@{}}{\footnotesize\textcolor{gray}{0.051}} \\ \textbf{0.051} \\ {\footnotesize\textcolor{gray}{0.051}}\end{tabular} & \begin{tabular}[c]{@{}c@{}}{\footnotesize\textcolor{gray}{3.974}} \\ 3.741 \\ {\footnotesize\textcolor{gray}{3.541}}\end{tabular} & \begin{tabular}[c]{@{}c@{}}{\footnotesize\textcolor{gray}{0.122}} \\ 0.120 \\ {\footnotesize\textcolor{gray}{0.117}}\end{tabular} & \begin{tabular}[c]{@{}c@{}}{\footnotesize\textcolor{gray}{0.101}} \\ 0.098 \\ {\footnotesize\textcolor{gray}{0.095}}\end{tabular} & \begin{tabular}[c]{@{}c@{}}{\footnotesize\textcolor{gray}{0.078}} \\ \textbf{0.077} \\ {\footnotesize\textcolor{gray}{0.076}}\end{tabular} & 3.7 & 1.449 \\
CatBoost & \begin{tabular}[c]{@{}c@{}}{\footnotesize\textcolor{gray}{98.480}} \\ 92.052 \\ {\footnotesize\textcolor{gray}{86.119}}\end{tabular} & \begin{tabular}[c]{@{}c@{}}{\footnotesize\textcolor{gray}{67.400}} \\ 61.351 \\ {\footnotesize\textcolor{gray}{56.008}}\end{tabular} & \begin{tabular}[c]{@{}c@{}}{\footnotesize\textcolor{gray}{0.011}} \\ 0.011 \\ {\footnotesize\textcolor{gray}{0.011}}\end{tabular} & \begin{tabular}[c]{@{}c@{}}{\footnotesize\textcolor{gray}{9.193}} \\ 9.022 \\ {\footnotesize\textcolor{gray}{8.811}}\end{tabular} & \begin{tabular}[c]{@{}c@{}}{\footnotesize\textcolor{gray}{0.051}} \\ \textbf{0.051} \\ {\footnotesize\textcolor{gray}{0.051}}\end{tabular} & \begin{tabular}[c]{@{}c@{}}{\footnotesize\textcolor{gray}{4.047}} \\ 3.839 \\ {\footnotesize\textcolor{gray}{3.652}}\end{tabular} & \begin{tabular}[c]{@{}c@{}}{\footnotesize\textcolor{gray}{0.110}} \\ 0.109 \\ {\footnotesize\textcolor{gray}{0.107}}\end{tabular} & \begin{tabular}[c]{@{}c@{}}{\footnotesize\textcolor{gray}{0.091}} \\ 0.087 \\ {\footnotesize\textcolor{gray}{0.085}}\end{tabular} & \begin{tabular}[c]{@{}c@{}}{\footnotesize\textcolor{gray}{0.081}} \\ 0.080 \\ {\footnotesize\textcolor{gray}{0.079}}\end{tabular} & 3.2 & 1.297 \\
M-TCN & \begin{tabular}[c]{@{}c@{}}{\footnotesize\textcolor{gray}{96.674}} \\ \textbf{90.942} \\ {\footnotesize\textcolor{gray}{85.371}}\end{tabular} & \begin{tabular}[c]{@{}c@{}}{\footnotesize\textcolor{gray}{65.837}} \\ \textbf{60.060} \\ {\footnotesize\textcolor{gray}{55.037}}\end{tabular} & \begin{tabular}[c]{@{}c@{}}{\footnotesize\textcolor{gray}{0.009}} \\ \textbf{0.009} \\ {\footnotesize\textcolor{gray}{0.009}}\end{tabular} & \begin{tabular}[c]{@{}c@{}}{\footnotesize\textcolor{gray}{2.484}} \\ \textbf{2.411} \\ {\footnotesize\textcolor{gray}{2.344}}\end{tabular} & \begin{tabular}[c]{@{}c@{}}{\footnotesize\textcolor{gray}{0.055}} \\ 0.055 \\ {\footnotesize\textcolor{gray}{0.055}}\end{tabular} & \begin{tabular}[c]{@{}c@{}}{\footnotesize\textcolor{gray}{2.518}} \\ \textbf{2.356} \\ {\footnotesize\textcolor{gray}{2.220}}\end{tabular} & \begin{tabular}[c]{@{}c@{}}{\footnotesize\textcolor{gray}{0.118}} \\ 0.116 \\ {\footnotesize\textcolor{gray}{0.115}}\end{tabular} & \begin{tabular}[c]{@{}c@{}}{\footnotesize\textcolor{gray}{0.076}} \\ \textbf{0.074} \\ {\footnotesize\textcolor{gray}{0.072}}\end{tabular} & \begin{tabular}[c]{@{}c@{}}{\footnotesize\textcolor{gray}{0.080}} \\ 0.079 \\ {\footnotesize\textcolor{gray}{0.078}}\end{tabular} & \textbf{2.2} & \textbf{1.027} \\
P-TSMixer & \begin{tabular}[c]{@{}c@{}}{\footnotesize\textcolor{gray}{109.455}} \\ 102.740 \\ {\footnotesize\textcolor{gray}{97.036}}\end{tabular} & \begin{tabular}[c]{@{}c@{}}{\footnotesize\textcolor{gray}{72.728}} \\ 67.104 \\ {\footnotesize\textcolor{gray}{61.907}}\end{tabular} & \begin{tabular}[c]{@{}c@{}}{\footnotesize\textcolor{gray}{0.033}} \\ 0.033 \\ {\footnotesize\textcolor{gray}{0.033}}\end{tabular} & \begin{tabular}[c]{@{}c@{}}{\footnotesize\textcolor{gray}{9.587}} \\ 9.321 \\ {\footnotesize\textcolor{gray}{9.024}}\end{tabular} & \begin{tabular}[c]{@{}c@{}}{\footnotesize\textcolor{gray}{0.054}} \\ 0.054 \\ {\footnotesize\textcolor{gray}{0.054}}\end{tabular} & \begin{tabular}[c]{@{}c@{}}{\footnotesize\textcolor{gray}{4.076}} \\ 3.876 \\ {\footnotesize\textcolor{gray}{3.716}}\end{tabular} & \begin{tabular}[c]{@{}c@{}}{\footnotesize\textcolor{gray}{0.276}} \\ 0.269 \\ {\footnotesize\textcolor{gray}{0.264}}\end{tabular} & \begin{tabular}[c]{@{}c@{}}{\footnotesize\textcolor{gray}{0.146}} \\ 0.142 \\ {\footnotesize\textcolor{gray}{0.140}}\end{tabular} & \begin{tabular}[c]{@{}c@{}}{\footnotesize\textcolor{gray}{0.150}} \\ 0.148 \\ {\footnotesize\textcolor{gray}{0.146}}\end{tabular} & 7.7 & 1.889 \\
xLSTM-M & \begin{tabular}[c]{@{}c@{}}{\footnotesize\textcolor{gray}{100.817}} \\ 94.353 \\ {\footnotesize\textcolor{gray}{88.400}}\end{tabular} & \begin{tabular}[c]{@{}c@{}}{\footnotesize\textcolor{gray}{70.090}} \\ 63.377 \\ {\footnotesize\textcolor{gray}{57.810}}\end{tabular} & \begin{tabular}[c]{@{}c@{}}{\footnotesize\textcolor{gray}{0.009}} \\ \textbf{0.009} \\ {\footnotesize\textcolor{gray}{0.009}}\end{tabular} & \begin{tabular}[c]{@{}c@{}}{\footnotesize\textcolor{gray}{7.632}} \\ 7.371 \\ {\footnotesize\textcolor{gray}{7.166}}\end{tabular} & \begin{tabular}[c]{@{}c@{}}{\footnotesize\textcolor{gray}{0.053}} \\ 0.052 \\ {\footnotesize\textcolor{gray}{0.052}}\end{tabular} & \begin{tabular}[c]{@{}c@{}}{\footnotesize\textcolor{gray}{4.126}} \\ 3.790 \\ {\footnotesize\textcolor{gray}{3.479}}\end{tabular} & \begin{tabular}[c]{@{}c@{}}{\footnotesize\textcolor{gray}{0.103}} \\ \textbf{0.101} \\ {\footnotesize\textcolor{gray}{0.100}}\end{tabular} & \begin{tabular}[c]{@{}c@{}}{\footnotesize\textcolor{gray}{0.077}} \\ 0.075 \\ {\footnotesize\textcolor{gray}{0.073}}\end{tabular} & \begin{tabular}[c]{@{}c@{}}{\footnotesize\textcolor{gray}{0.082}} \\ 0.081 \\ {\footnotesize\textcolor{gray}{0.080}}\end{tabular} & 3.3 & \underline{1.222} \\
TST & \begin{tabular}[c]{@{}c@{}}{\footnotesize\textcolor{gray}{103.926}} \\ 97.340 \\ {\footnotesize\textcolor{gray}{91.163}}\end{tabular} & \begin{tabular}[c]{@{}c@{}}{\footnotesize\textcolor{gray}{70.092}} \\ 63.627 \\ {\footnotesize\textcolor{gray}{58.515}}\end{tabular} & \begin{tabular}[c]{@{}c@{}}{\footnotesize\textcolor{gray}{0.017}} \\ 0.017 \\ {\footnotesize\textcolor{gray}{0.017}}\end{tabular} & \begin{tabular}[c]{@{}c@{}}{\footnotesize\textcolor{gray}{7.035}} \\ 6.889 \\ {\footnotesize\textcolor{gray}{6.748}}\end{tabular} & \begin{tabular}[c]{@{}c@{}}{\footnotesize\textcolor{gray}{0.054}} \\ 0.054 \\ {\footnotesize\textcolor{gray}{0.054}}\end{tabular} & \begin{tabular}[c]{@{}c@{}}{\footnotesize\textcolor{gray}{4.023}} \\ 3.789 \\ {\footnotesize\textcolor{gray}{3.561}}\end{tabular} & \begin{tabular}[c]{@{}c@{}}{\footnotesize\textcolor{gray}{0.113}} \\ 0.112 \\ {\footnotesize\textcolor{gray}{0.110}}\end{tabular} & \begin{tabular}[c]{@{}c@{}}{\footnotesize\textcolor{gray}{0.080}} \\ 0.078 \\ {\footnotesize\textcolor{gray}{0.076}}\end{tabular} & \begin{tabular}[c]{@{}c@{}}{\footnotesize\textcolor{gray}{0.083}} \\ 0.082 \\ {\footnotesize\textcolor{gray}{0.081}}\end{tabular} & 4.4 & 1.329 \\
\midrule
TST-PGD & \begin{tabular}[c]{@{}c@{}}{\footnotesize\textcolor{gray}{102.189}} \\ 94.721 \\ {\footnotesize\textcolor{gray}{88.588}}\end{tabular} & \begin{tabular}[c]{@{}c@{}}{\footnotesize\textcolor{gray}{67.419}} \\ 61.901 \\ {\footnotesize\textcolor{gray}{57.215}}\end{tabular} & \begin{tabular}[c]{@{}c@{}}{\footnotesize\textcolor{gray}{0.024}} \\ 0.024 \\ {\footnotesize\textcolor{gray}{0.024}}\end{tabular} & \begin{tabular}[c]{@{}c@{}}{\footnotesize\textcolor{gray}{13.264}} \\ 13.069 \\ {\footnotesize\textcolor{gray}{12.852}}\end{tabular} & \begin{tabular}[c]{@{}c@{}}{\footnotesize\textcolor{gray}{0.055}} \\ 0.054 \\ {\footnotesize\textcolor{gray}{0.054}}\end{tabular} & \begin{tabular}[c]{@{}c@{}}{\footnotesize\textcolor{gray}{3.669}} \\ 3.455 \\ {\footnotesize\textcolor{gray}{3.246}}\end{tabular} & \begin{tabular}[c]{@{}c@{}}{\footnotesize\textcolor{gray}{0.211}} \\ 0.207 \\ {\footnotesize\textcolor{gray}{0.204}}\end{tabular} & \begin{tabular}[c]{@{}c@{}}{\footnotesize\textcolor{gray}{0.140}} \\ 0.136 \\ {\footnotesize\textcolor{gray}{0.132}}\end{tabular} & \begin{tabular}[c]{@{}c@{}}{\footnotesize\textcolor{gray}{0.141}} \\ 0.138 \\ {\footnotesize\textcolor{gray}{0.136}}\end{tabular} & 6.2 & 1.764 {\scriptsize\textcolor{red}{$\uparrow$+0.435}} \\
TST-ISensD & \begin{tabular}[c]{@{}c@{}}{\footnotesize\textcolor{gray}{102.909}} \\ 95.732 \\ {\footnotesize\textcolor{gray}{89.714}}\end{tabular} & \begin{tabular}[c]{@{}c@{}}{\footnotesize\textcolor{gray}{72.737}} \\ 67.344 \\ {\footnotesize\textcolor{gray}{62.294}}\end{tabular} & \begin{tabular}[c]{@{}c@{}}{\footnotesize\textcolor{gray}{0.026}} \\ 0.026 \\ {\footnotesize\textcolor{gray}{0.025}}\end{tabular} & \begin{tabular}[c]{@{}c@{}}{\footnotesize\textcolor{gray}{12.238}} \\ 12.052 \\ {\footnotesize\textcolor{gray}{11.852}}\end{tabular} & \begin{tabular}[c]{@{}c@{}}{\footnotesize\textcolor{gray}{0.054}} \\ 0.054 \\ {\footnotesize\textcolor{gray}{0.054}}\end{tabular} & \begin{tabular}[c]{@{}c@{}}{\footnotesize\textcolor{gray}{4.887}} \\ 4.654 \\ {\footnotesize\textcolor{gray}{4.401}}\end{tabular} & \begin{tabular}[c]{@{}c@{}}{\footnotesize\textcolor{gray}{0.159}} \\ 0.156 \\ {\footnotesize\textcolor{gray}{0.153}}\end{tabular} & \begin{tabular}[c]{@{}c@{}}{\footnotesize\textcolor{gray}{0.110}} \\ 0.107 \\ {\footnotesize\textcolor{gray}{0.105}}\end{tabular} & \begin{tabular}[c]{@{}c@{}}{\footnotesize\textcolor{gray}{0.109}} \\ 0.108 \\ {\footnotesize\textcolor{gray}{0.106}}\end{tabular} & 7.3 & 1.687 {\scriptsize\textcolor{red}{$\uparrow$+0.358}} \\
TST-F2F & \begin{tabular}[c]{@{}c@{}}{\footnotesize\textcolor{gray}{117.895}} \\ 108.129 \\ {\footnotesize\textcolor{gray}{100.705}}\end{tabular} & \begin{tabular}[c]{@{}c@{}}{\footnotesize\textcolor{gray}{68.616}} \\ 62.415 \\ {\footnotesize\textcolor{gray}{58.105}}\end{tabular} & \begin{tabular}[c]{@{}c@{}}{\footnotesize\textcolor{gray}{0.012}} \\ 0.012 \\ {\footnotesize\textcolor{gray}{0.012}}\end{tabular} & \begin{tabular}[c]{@{}c@{}}{\footnotesize\textcolor{gray}{15.820}} \\ 15.471 \\ {\footnotesize\textcolor{gray}{15.151}}\end{tabular} & \begin{tabular}[c]{@{}c@{}}{\footnotesize\textcolor{gray}{0.061}} \\ 0.061 \\ {\footnotesize\textcolor{gray}{0.060}}\end{tabular} & \begin{tabular}[c]{@{}c@{}}{\footnotesize\textcolor{gray}{4.545}} \\ 4.283 \\ {\footnotesize\textcolor{gray}{4.042}}\end{tabular} & \begin{tabular}[c]{@{}c@{}}{\footnotesize\textcolor{gray}{0.151}} \\ 0.148 \\ {\footnotesize\textcolor{gray}{0.144}}\end{tabular} & \begin{tabular}[c]{@{}c@{}}{\footnotesize\textcolor{gray}{0.110}} \\ 0.107 \\ {\footnotesize\textcolor{gray}{0.104}}\end{tabular} & \begin{tabular}[c]{@{}c@{}}{\footnotesize\textcolor{gray}{0.110}} \\ 0.108 \\ {\footnotesize\textcolor{gray}{0.106}}\end{tabular} & 6.9 & 1.601 {\scriptsize\textcolor{red}{$\uparrow$+0.272}} \\
\bottomrule
\end{tabular}
}
\end{center}
\end{table*}

\section{Normalized performance under corruption (nPC)}
\label{app:npc}

Dividing mPC by the RMSE of the na\"{i}ve mean
predictor, a severity-invariant baseline that
always predicts the training-set mean, yields:
\begin{equation}\label{eq:npc}
  \mathrm{nPC}
  = \frac{\mathrm{mPC}}{\mathrm{RMSE}_{\mathrm{naive}}}\,.
\end{equation}

A value below~$1$ indicates that the model adds value over this trivial
reference; a value of 1 marks the point at which the model is, on average, no better than the naïve predictor. The omnibus test is
not significant (see Appendix~\ref{app:significance}), so
results should be read descriptively.

\begin{table*}[h]
\caption{nPC per model and dataset. Bold = best, \underline{underline} = second-best per column. GM is the geometric mean across datasets; arrows show GM change vs. TST.}
\label{tab:npc}
\begin{center}
\setlength{\tabcolsep}{1.5pt}
\adjustbox{max width=\textwidth}{%
\begin{tabular}{l c c c c c c c c c c l}
\toprule
\textbf{Model} & \textbf{PM10} & \textbf{PM2.5} & \textbf{Batt.} & \textbf{HR} & \textbf{TEP} & \textbf{Veh.} & \textbf{MMR} & \textbf{T66-13} & \textbf{T66-14} & \textbf{Rank} & \textbf{GM} \\
\midrule
XGBoost & \textbf{0.712} & \underline{0.587} & 0.227 & 0.543 & \underline{0.768} & \textbf{0.537} & \underline{0.259} & \underline{0.324} & \underline{0.267} & \textbf{2.3} & \underline{0.428} \\
CatBoost & 0.724 & 0.597 & \underline{0.172} & 0.503 & \textbf{0.767} & 0.604 & \textbf{0.256} & \textbf{0.303} & \textbf{0.262} & 2.6 & \textbf{0.413} \\
M-TCN & 0.724 & 0.611 & 0.345 & \textbf{0.231} & 0.876 & 1.378 & 0.334 & 0.374 & 0.362 & 5.6 & 0.499 \\
P-TSMixer & 0.785 & 0.658 & 0.358 & 0.449 & 0.814 & 0.958 & 0.522 & 0.589 & 0.559 & 7.2 & 0.607 \\
xLSTM-M & \underline{0.721} & 0.595 & 0.215 & \underline{0.423} & 0.789 & \underline{0.540} & 0.280 & 0.339 & 0.319 & 3.0 & 0.430 \\
TST & 0.779 & 0.651 & 0.270 & 0.549 & 0.810 & 0.988 & 0.330 & 0.395 & 0.387 & 6.0 & 0.525 \\
\midrule
TST-PGD & 0.743 & \textbf{0.584} & 0.285 & 0.674 & 0.819 & 0.553 & 0.381 & 0.414 & 0.384 & 5.6 & 0.509 {\scriptsize\textcolor{ForestGreen}{$\downarrow$-0.016}} \\
TST-ISensD & 0.796 & 0.659 & 0.443 & 0.673 & 0.819 & 0.685 & 0.370 & 0.474 & 0.452 & 7.7 & 0.576 {\scriptsize\textcolor{red}{$\uparrow$+0.051}} \\
TST-F2F & 0.868 & 0.608 & \textbf{0.113} & 1.032 & 1.115 & 0.600 & 0.276 & 0.331 & 0.290 & 5.1 & 0.469 {\scriptsize\textcolor{ForestGreen}{$\downarrow$-0.056}} \\
\bottomrule
\end{tabular}
}
\end{center}
\end{table*}

\section{Metric discussion}
\label{app:metric_discussion}

\paragraph{Relationship between rPC and nPC.}
The two normalized metrics are linked by a model-specific scaling factor:
\begin{equation}\label{eq:link}
  \mathrm{rPC} = \mathrm{nPC} \times
  \frac{\mathrm{RMSE}_{\mathrm{naive}}}{\mathrm{RMSE}_{\mathrm{clean}}}\,.
\end{equation}
The ratio
$\mathrm{RMSE}_{\mathrm{naive}}/\mathrm{RMSE}_{\mathrm{clean}}$
is itself interpretable: it measures how much better than na\"{i}ve the
model is on clean data, and thus quantifies the headroom available for
degradation before the model becomes useless.

\paragraph{Relationship to mCE.}
The normalized Performance under Corruption (nPC) is analogous to the mean
Corruption Error (mCE) of
ImageNet-C~\cite{hendrycks_benchmarking_2018}, with one deliberate
departure in the choice of reference.
In the original ImageNet-C protocol, each corruption type's summed error is
normalized by AlexNet's error on that same corruption, equalizing the
difficulty of different corruption types in the denominator.
nPC instead uses the na\"{i}ve mean predictor, a severity-invariant
baseline, as a single, shared reference.
Because this denominator is constant across models, severities, and failure
modes, differences in nPC directly reflect the intrinsic difficulty of each
failure mode and dataset, which we consider informative for practitioners
assessing deployment risk.
This choice also means that nPC preserves difficulty differences across
failure modes as a finding of interest, rather than normalizing them away.

\paragraph{Relationship between rPC, Relative mCE, and the Robustness Score.}
The relative Performance under Corruption (rPC) was introduced by
Michaelis~et~al.~\cite{michaelis_benchmarking_2020} to isolate degradation
from nominal accuracy by normalizing with the model's own clean-data
performance.
Hendrycks and Dietterich~\cite{hendrycks_benchmarking_2018} achieve the
same goal with the Relative~mCE, which \emph{subtracts} the clean error
rate before normalizing by AlexNet. Windmann~et~al.~\cite{windmann_quantifying_2025} define a closely related
quantity, the relative performance ratio
$u_{\mathrm{rel}} = \mathrm{RMSE}_{\mathrm{clean}} /
\mathrm{RMSE}_{d,c,s}$,
which is simply the reciprocal of the per-combination degradation ratio.
We use the rPC that shares the ``lower-is-better''
direction of mPC, making tables and rankings immediately readable without
direction-switching.

\section{Per-failure-mode evaluation}
\label{app:per_fm}

The main evaluation aggregates over
all failure modes, features, and severities into a single mPC and rPC
per model and dataset.
Here we report both metrics \emph{per failure mode}, retaining the
average over features and severities but disaggregating across failure
modes to reveal which corruption types drive the overall scores.
 
\paragraph{Computation.}
For a given failure mode~$d \in \mathcal{D}$, we restrict the
summation in Equation~\eqref{eq:mpc} to that mode and define
\begin{equation}\label{eq:mpc_fm}
  \mathrm{mPC}_{d}
  = \frac{1}{|\mathcal{C}|\,|\mathcal{S}|}
    \sum_{c \in \mathcal{C}}\sum_{s \in \mathcal{S}}
    \mathrm{RMSE}_{d,c,s}\,,
\end{equation}
the mean RMSE under failure mode~$d$, averaged over all affected
features~$c$ and severities~$s$.
The per-failure-mode relative metric follows Equation~\eqref{eq:rpc}
directly:
\begin{equation}\label{eq:rpc_fm}
  \mathrm{rPC}_{d}
  = \frac{\mathrm{mPC}_{d}}{\mathrm{RMSE}_{\mathrm{clean}}}\,.
\end{equation}
 
\paragraph{Cross-dataset aggregation.}
We aggregate across datasets using the same protocol as in the main
paper (Section~\ref{sec:eval_protocol}).
$\mathrm{rPC}_{d}$ is a dimensionless ratio and enters the geometric
mean directly.
$\mathrm{mPC}_{d}$ carries dataset-specific units; for each
dataset--failure-mode pair we first normalize by the best model
($1.0$~=~best), then take the geometric mean across datasets.
 
\paragraph{Failure-mode difficulty.}
To rank failure modes by difficulty we compute, for each failure
mode~$d$, the geometric mean across all models, of $\mathrm{rPC}_{d}$
for the relative view and of the normalized $\mathrm{mPC}_{d}$ for the
absolute view.
A higher value indicates that the failure mode causes larger
degradation on average; the rPC-based ranking is independent of nominal
accuracy, while the mPC-based ranking additionally reflects absolute
error magnitude.
Tables~\ref{tab:mpc_fm_sorted} and~\ref{tab:rpc_fm_sorted} sort rows by the
respective difficulty score in ascending order (easiest to hardest).
 
\paragraph{Analysis.}
The per-failure-mode breakdown reveals substantial heterogeneity in
corruption difficulty.
The easiest failure modes increase rPC only marginally above~$1$,
whereas the hardest modes raise it considerably, motivating the
finer-grained view provided here as a complement to the aggregate
scores in the main paper.
Model rankings also shift across failure modes: a model that leads
under one corruption type may be among the most affected under another,
highlighting the practical value of per-failure-mode reporting for
practitioners selecting models for deployment environments with known
sensor characteristics.
Full per-failure-mode tables are provided in
Tables~\ref{tab:mpc_fm_sorted}~and~\ref{tab:rpc_fm_sorted}.

\begin{table*}[h]
\caption{Per-failure-mode normalized mPC. Rows sorted by difficulty (top = easiest). Bold = best model per failure mode; green/red cells = each model's best/worst failure mode; shaded columns are robust TST variants. GM = geometric mean across models.}
\label{tab:mpc_fm_sorted}
\begin{center}
\setlength{\tabcolsep}{1.5pt}
\adjustbox{max width=\textwidth}{%
\begin{tabular}{l c c c c c c >{\columncolor{gray!10}}l >{\columncolor{gray!10}}l >{\columncolor{gray!10}}l | c}
\toprule
\textbf{Failure} & \textbf{XGBoost} & \textbf{CatBoost} & \textbf{M-TCN} & \textbf{P-TSMixer} & \textbf{xLSTM-M} & \textbf{TST} & \textbf{TST-PGD} & \textbf{TST-ISensD} & \textbf{TST-F2F} & \textbf{GM} \\
\midrule
Trim. (c) & 1.248 & 1.173 & \textbf{1.076} & 1.666 & 1.140 & 1.250 & 1.511 & \cellcolor{green!20}1.360 & 1.366 & 1.299 \\
Trim. (v) & 1.301 & 1.200 & \textbf{\cellcolor{green!20}1.035} & 1.691 & 1.153 & \cellcolor{green!20}1.244 & 1.574 & 1.471 & 1.469 & 1.333 \\
Outl. & 1.227 & 1.128 & \textbf{1.057} & 1.664 & 1.114 & 1.418 & \cellcolor{red!20}1.656 & 1.833 & 1.437 & 1.368 \\
TV Scal. & 1.216 & \textbf{1.148} & 1.303 & 1.670 & 1.224 & 1.367 & 1.460 & 1.635 & 1.424 & 1.372 \\
Scal. & 1.174 & \textbf{1.133} & 1.584 & 1.684 & 1.176 & 1.481 & 1.428 & 1.670 & 1.500 & 1.411 \\
Noise & \cellcolor{green!20}1.149 & \cellcolor{green!20}1.100 & 1.518 & \cellcolor{green!20}1.599 & \textbf{\cellcolor{green!20}1.033} & 1.545 & 1.503 & \cellcolor{red!20}2.016 & \cellcolor{red!20}1.553 & 1.417 \\
Hard F. & 1.221 & 1.232 & 1.713 & 1.830 & 1.284 & \cellcolor{red!20}1.731 & \cellcolor{green!20}1.372 & 1.500 & \textbf{\cellcolor{green!20}1.202} & 1.436 \\
NL Dr. & \cellcolor{red!20}1.349 & \cellcolor{red!20}1.290 & 1.331 & 1.830 & \cellcolor{red!20}1.444 & 1.550 & 1.496 & 1.691 & \textbf{1.285} & 1.464 \\
Lin. Dr. & 1.319 & 1.287 & 1.438 & 1.849 & 1.414 & 1.592 & 1.455 & 1.685 & \textbf{1.258} & 1.467 \\
Bias & 1.273 & 1.270 & \cellcolor{red!20}1.736 & \cellcolor{red!20}1.864 & 1.343 & 1.717 & 1.430 & 1.677 & \textbf{1.241} & 1.489 \\
\bottomrule
\end{tabular}
}
\end{center}
\end{table*}

\begin{table*}[h]
\caption{Per-failure-mode rPC. Rows sorted by difficulty (top = easiest). Bold = best model per failure mode; green/red cells = each model's best/worst failure mode; shaded columns are robust TST variants. GM = geometric mean of rPC across models.}
\label{tab:rpc_fm_sorted}
\begin{center}
\setlength{\tabcolsep}{1.5pt}
\adjustbox{max width=\textwidth}{%
\begin{tabular}{l c c c c c c >{\columncolor{gray!10}}l >{\columncolor{gray!10}}l >{\columncolor{gray!10}}l | c}
\toprule
\textbf{Failure} & \textbf{XGBoost} & \textbf{CatBoost} & \textbf{M-TCN} & \textbf{P-TSMixer} & \textbf{xLSTM-M} & \textbf{TST} & \textbf{TST-PGD} & \textbf{TST-ISensD} & \textbf{TST-F2F} & \textbf{GM} \\
\midrule
Trim. (v) & \cellcolor{green!20}1.047 & \cellcolor{green!20}1.080 & \cellcolor{green!20}1.177 & \cellcolor{green!20}1.044 & \cellcolor{green!20}1.100 & \cellcolor{green!20}1.092 & \cellcolor{green!20}1.041 & \textbf{\cellcolor{green!20}1.017} & \cellcolor{green!20}1.070 & 1.073 \\
Trim. (c) & 1.089 & 1.144 & 1.327 & 1.116 & 1.180 & 1.190 & 1.084 & \textbf{1.020} & 1.079 & 1.134 \\
Outl. & \textbf{1.067} & 1.096 & 1.298 & 1.110 & 1.148 & 1.344 & 1.183 & 1.368 & 1.131 & 1.189 \\
TV Scal. & 1.143 & 1.206 & 1.729 & 1.204 & 1.364 & 1.400 & \textbf{1.127} & 1.319 & 1.211 & 1.289 \\
NL Dr. & 1.338 & 1.430 & 1.863 & 1.392 & 1.698 & 1.675 & 1.219 & 1.440 & \textbf{1.153} & 1.451 \\
Scal. & 1.216 & 1.311 & 2.315 & 1.337 & 1.443 & 1.671 & \textbf{1.215} & 1.484 & 1.405 & 1.460 \\
Noise & \textbf{1.255} & 1.342 & 2.339 & 1.339 & 1.336 & 1.839 & 1.348 & \cellcolor{red!20}1.890 & \cellcolor{red!20}1.534 & 1.547 \\
Lin. Dr. & 1.396 & 1.522 & 2.148 & 1.501 & 1.774 & 1.836 & 1.265 & 1.531 & \textbf{1.205} & 1.552 \\
Bias & 1.548 & 1.725 & 2.978 & 1.738 & 1.935 & 2.275 & 1.428 & 1.750 & \textbf{1.365} & 1.809 \\
Hard F. & \cellcolor{red!20}1.634 & \cellcolor{red!20}1.843 & \cellcolor{red!20}3.238 & \cellcolor{red!20}1.879 & \cellcolor{red!20}2.038 & \cellcolor{red!20}2.526 & \cellcolor{red!20}1.509 & 1.725 & \textbf{1.456} & 1.922 \\
\bottomrule
\end{tabular}
}
\end{center}
\end{table*}

\section{Mean imputation as a failure mitigation strategy}
\label{app:isensd_discussion}

The main benchmark evaluates robustness across a wide range of failure
modes, severities, and affected channels.
In this section we analyze a practically relevant special case:
a deployment scenario in which (i)~a reliable failure detection
mechanism is available, and (ii)~the operator can intervene on the
input signal.
When a sensor is detected as faulty, the simplest corrective action is
to replace the failing channel with its training-set mean, which, under
$z$-score normalization, amounts to setting it to zero.
This corresponds to the \emph{hard-fault} failure mode at
severity $s\!=\!0$ in our benchmark.
 
\paragraph{Metrics.}
Since both the failure mode and severity are fixed, the mPC and rPC
definitions (Eqs.~\eqref{eq:mpc}--\eqref{eq:rpc}) reduce to an
average over affected channels only:
\begin{equation}\label{eq:mpc_hf0}
  \mathrm{mPC}_{\mathrm{hf},0}
  = \frac{1}{|\mathcal{C}|}\sum_{c \in \mathcal{C}}
    \mathrm{RMSE}_{\mathrm{hf},\,c,\,0}\,,
\end{equation}
\begin{equation}\label{eq:rpc_hf0}
  \mathrm{rPC}_{\mathrm{hf},0}
  = \frac{\mathrm{mPC}_{\mathrm{hf},0}}
         {\mathrm{RMSE}_{\mathrm{clean}}}\,.
\end{equation}
Cross-dataset aggregation follows the protocol in
Section~\ref{sec:eval_protocol}: $\mathrm{rPC}_{\mathrm{hf},0}$ enters
the geometric mean directly; $\mathrm{mPC}_{\mathrm{hf},0}$ is first
normalized per dataset by the best model before computing the geometric
mean.
 
\paragraph{Results.}
Tables~\ref{tab:mpc_hf0} and~\ref{tab:rpc_hf0} report mPC and rPC for
all models under this scenario.
TST-ISensD achieves the best geometric-mean rPC ($1.034$) and mPC
($1.072$) as well as the best mean rank on both metrics, and on rPC
significantly outranks CatBoost, xLSTM-M, TST, and M-TCN
(Appendix~\ref{app:significance}).
On rPC, TST-ISensD attains near-perfect scores close to $1.0$ on every
dataset, indicating virtually no degradation when a channel is
replaced by its mean, a direct consequence of its training procedure,
which randomly drops input channels by setting them to zero.
The model has explicitly learned to
produce accurate predictions when any single channel is absent.  Among
the remaining models, TST-F2F (GM rPC $1.208$) and TST-PGD ($1.275$)
also improve substantially over the base TST ($1.718$), though neither
approaches ISensD's consistency.
 
\paragraph{Practical implication.}
These results suggest that when reliable failure detection and input
intervention are available, ISensD, a simple dropout-based
augmentation applied to input channels during training, provides an
effective and easy-to-implement strategy.
It transforms a scenario that degrades most models into one where
prediction quality is on average nearly unaffected, without requiring
architectural changes or adversarial training.

\begin{table*}[h]
\caption{mPC under mean imputation (hard-fault at severity s=0); Bold/\underline{underline}: best/second-best per column. GM is the geometric mean of best-normalized values (1.0 = best); arrows show GM change vs. TST.}
\label{tab:mpc_hf0}
\begin{center}
\setlength{\tabcolsep}{1.5pt}
\adjustbox{max width=\textwidth}{%
\begin{tabular}{l c c c c c c c c c c l}
\toprule
\textbf{Model} & \textbf{PM10} & \textbf{PM2.5} & \textbf{Batt.} & \textbf{HR} & \textbf{TEP} & \textbf{Veh.} & \textbf{MMR} & \textbf{T66-13} & \textbf{T66-14} & \textbf{Rank} & \textbf{GM} \\
\midrule
XGBoost & \underline{98.61} & 68.43 & 0.07 & 13.58 & \textbf{0.05} & 4.82 & 0.18 & \underline{0.12} & \textbf{0.10} & 3.9 & 1.209 \\
CatBoost & 101.66 & 68.81 & \underline{0.06} & 12.92 & \textbf{0.05} & 4.83 & \underline{0.17} & \textbf{0.11} & 0.12 & 3.4 & 1.195 \\
M-TCN & 99.15 & 71.02 & 0.07 & \textbf{8.18} & \underline{0.06} & 11.17 & 0.19 & \underline{0.12} & 0.13 & 5.6 & 1.304 \\
P-TSMixer & 106.20 & 73.66 & 0.10 & \underline{12.22} & \textbf{0.05} & 7.09 & 0.34 & 0.23 & 0.23 & 7.2 & 1.658 \\
xLSTM-M & 101.45 & 71.94 & \underline{0.06} & 13.00 & \textbf{0.05} & \underline{4.65} & 0.18 & \textbf{0.11} & 0.12 & 3.9 & 1.208 \\
TST & 108.01 & 76.68 & \underline{0.06} & 15.12 & \textbf{0.05} & 9.03 & 0.20 & 0.13 & 0.14 & 6.8 & 1.403 \\
\midrule
TST-PGD & 112.96 & \textbf{67.15} & 0.07 & 16.54 & \textbf{0.05} & 5.37 & 0.23 & 0.15 & 0.15 & 6.8 & 1.381 {\scriptsize\textcolor{ForestGreen}{$\downarrow$-0.021}} \\
TST-ISensD & \textbf{97.07} & \underline{67.90} & \textbf{0.03} & 13.21 & \textbf{0.05} & \textbf{4.62} & \textbf{0.16} & \textbf{0.11} & \underline{0.11} & \textbf{2.2} & \textbf{1.072} {\scriptsize\textcolor{ForestGreen}{$\downarrow$-0.331}} \\
TST-F2F & 114.68 & 70.63 & \textbf{0.03} & 17.61 & 0.07 & 4.98 & \textbf{0.16} & \underline{0.12} & \underline{0.11} & 5.2 & \underline{1.188} {\scriptsize\textcolor{ForestGreen}{$\downarrow$-0.215}} \\
\bottomrule
\end{tabular}
}
\end{center}
\end{table*}

\begin{table*}[h]
\caption{rPC under mean imputation (hard-fault at severity s=0); Bold/\underline{underline}: best/second-best per column. GM is the geometric mean across datasets; arrows show GM change vs. TST.}
\label{tab:rpc_hf0}
\begin{center}
\setlength{\tabcolsep}{1.5pt}
\adjustbox{max width=\textwidth}{%
\begin{tabular}{l c c c c c c c c c c l}
\toprule
\textbf{Model} & \textbf{PM10} & \textbf{PM2.5} & \textbf{Batt.} & \textbf{HR} & \textbf{TEP} & \textbf{Veh.} & \textbf{MMR} & \textbf{T66-13} & \textbf{T66-14} & \textbf{Rank} & \textbf{GM} \\
\midrule
XGBoost & 1.076 & 1.121 & 2.855 & 1.428 & 1.007 & 1.289 & 1.482 & 1.257 & 1.319 & 4.8 & 1.358 \\
CatBoost & 1.104 & 1.122 & 5.602 & 1.432 & 1.007 & 1.258 & 1.596 & 1.306 & 1.464 & 5.7 & 1.500 \\
M-TCN & 1.090 & 1.182 & 7.315 & 3.393 & 1.043 & 4.741 & 1.677 & 1.601 & 1.628 & 8.0 & 2.068 \\
P-TSMixer & \underline{1.034} & 1.098 & 2.988 & 1.311 & 1.004 & 1.830 & 1.249 & 1.594 & 1.525 & 4.6 & 1.429 \\
xLSTM-M & 1.075 & 1.135 & 6.658 & 1.764 & 1.006 & 1.228 & 1.758 & 1.505 & 1.539 & 6.2 & 1.609 \\
TST & 1.110 & 1.205 & 3.685 & 2.195 & 1.005 & 2.383 & 1.749 & 1.667 & 1.727 & 7.7 & 1.718 \\
\midrule
TST-PGD & 1.193 & \underline{1.085} & 2.728 & 1.265 & \underline{1.003} & 1.553 & \underline{1.098} & \underline{1.069} & 1.090 & 3.6 & 1.275 {\scriptsize\textcolor{ForestGreen}{$\downarrow$-0.443}} \\
TST-ISensD & \textbf{1.014} & \textbf{1.008} & \textbf{1.160} & \textbf{1.096} & \textbf{0.999} & \textbf{0.992} & \textbf{1.016} & \textbf{1.014} & \textbf{1.021} & \textbf{1.0} & \textbf{1.034} {\scriptsize\textcolor{ForestGreen}{$\downarrow$-0.684}} \\
TST-F2F & 1.061 & 1.132 & \underline{2.405} & \underline{1.138} & 1.126 & \underline{1.163} & 1.109 & 1.080 & \underline{1.064} & 3.6 & \underline{1.208} {\scriptsize\textcolor{ForestGreen}{$\downarrow$-0.510}} \\
\bottomrule
\end{tabular}
}
\end{center}
\end{table*}

\newpage

\section{Severity curves}
\label{app:sev}

\begin{figure}[h]
    \centering
    \includegraphics[width=0.9\linewidth]{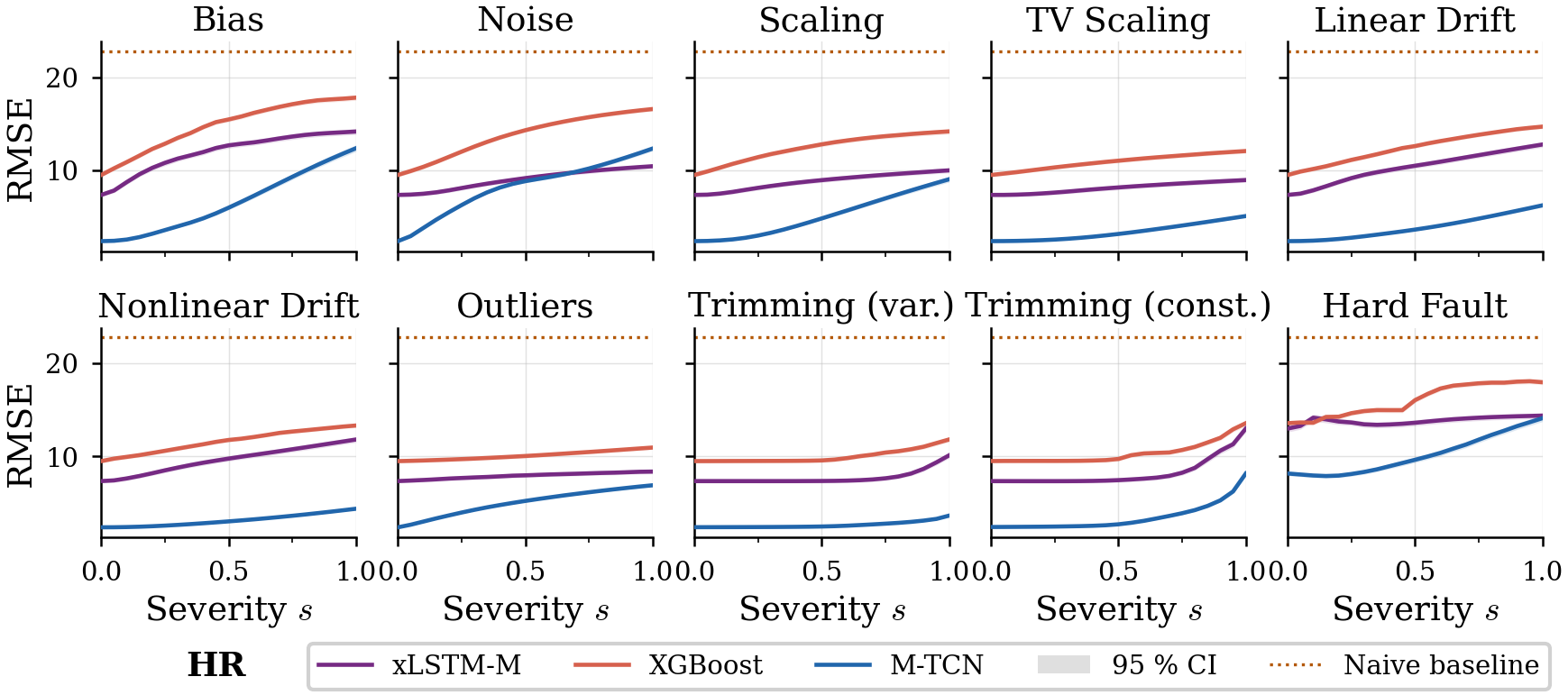}
     \caption{HR severity curves: XGBoost, xLSTM-M, M-TCN.}
    \label{fig:sev_ppg}
    \vspace{-10pt}
\end{figure}

\begin{figure}[h]
    \centering
    \includegraphics[width=0.9\linewidth]{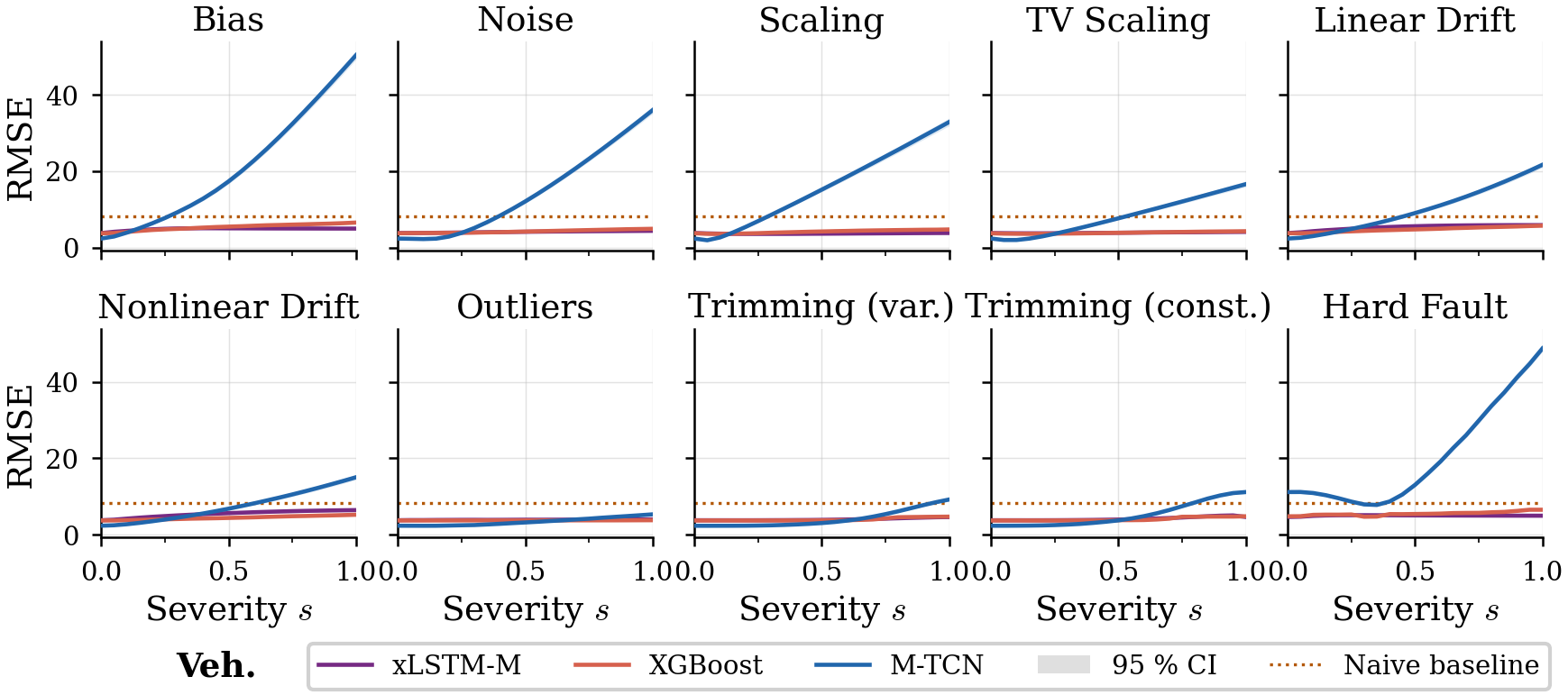}
     \caption{Veh. severity curves: XGBoost, xLSTM-M, M-TCN.}
    \label{fig:sev_veh}
    \vspace{-10pt}
\end{figure}

\begin{figure}[h]
    \centering
    \includegraphics[width=0.9\linewidth]{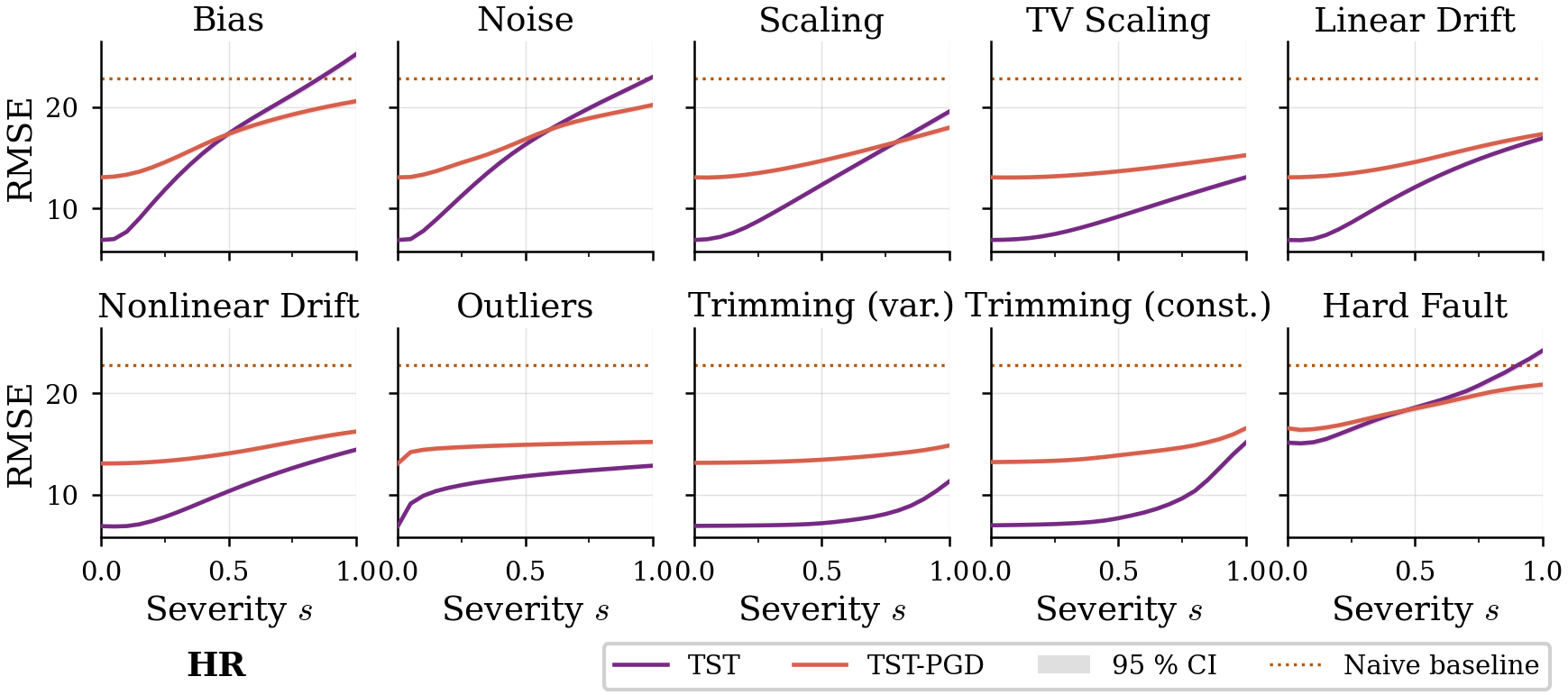}
     \caption{HR severity curves: TST vs. TST-PGD.}
    \label{fig:sev_ppg_rob}
    \vspace{-10pt}
\end{figure}

\begin{figure}[h]
    \centering
    \includegraphics[width=0.9\linewidth]{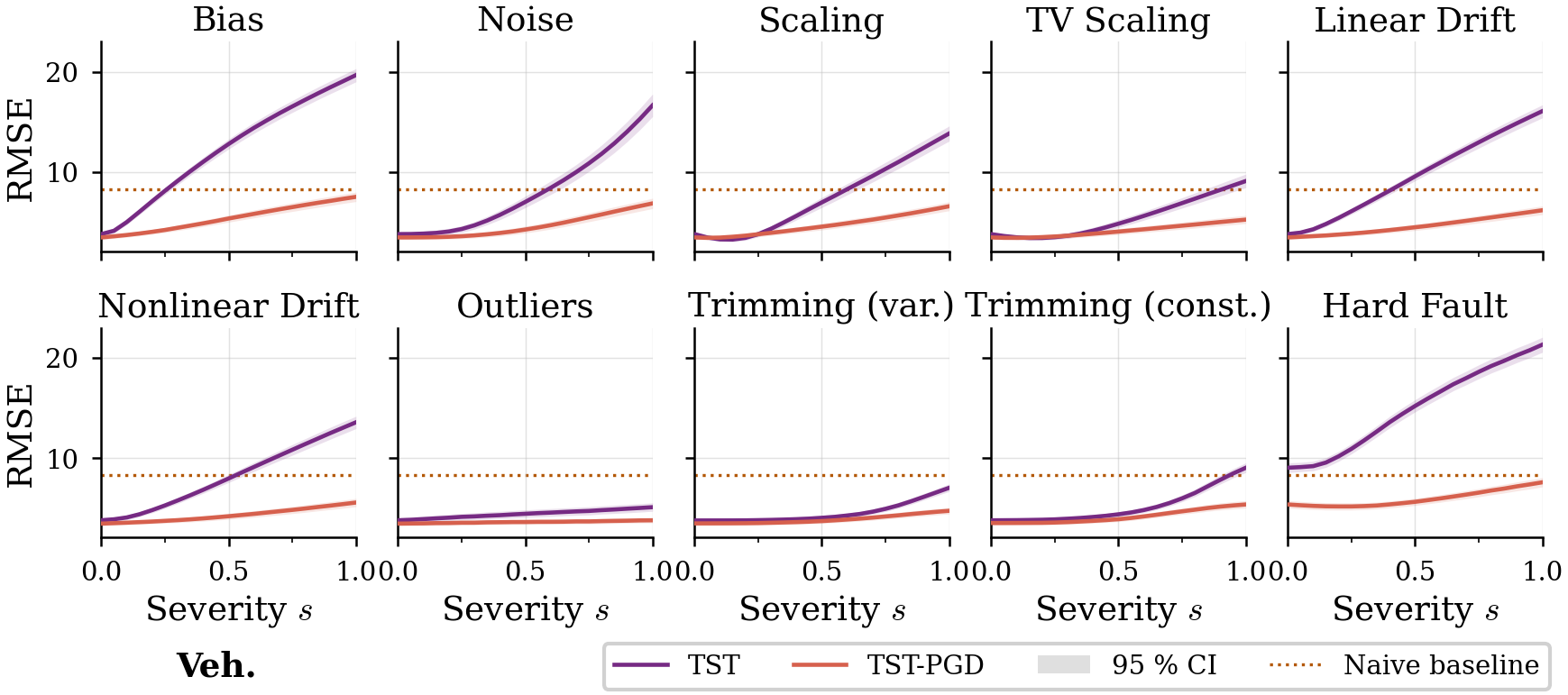}
     \caption{Veh. severity curves: TST vs. TST-PGD.}
    \label{fig:sev_veh_rob}
    \vspace{-10pt}
\end{figure}

The aggregate scores in the main paper compress an entire severity sweep 
into a single number per (model, dataset) pair. To give the reader a sense 
of the underlying behaviour, we show four illustrative slices of the full 
sweep: per-failure-mode severity curves on two datasets, heartrate estimation (HR) 
and tire temperature estimation (Veh.), for two model groupings. Figures~\ref{fig:sev_ppg} 
and~\ref{fig:sev_veh} compare three non-robustified baselines (XGBoost, 
xLSTM-M, M-TCN); Figures~\ref{fig:sev_ppg_rob} and~\ref{fig:sev_veh_rob} compare vanilla TST 
against its adversarially-trained variant TST-PGD.

\paragraph{Baseline behaviour across datasets.}
On HR (Figure~\ref{fig:sev_ppg}), M-TCN starts from a 
substantially lower clean RMSE than xLSTM-M or XGBoost, reflecting its 
strong nominal performance (Appendix~\ref{app:nom}). Although M-TCN degrades faster in relative terms, 
its absolute corrupted RMSE remains lowest across most failure modes, which 
is exactly what mPC rewards: M-TCN is the best mPC model on HR 
(Table~\ref{tab:mpc}). On Vehicle Dynamics (Figure~\ref{fig:sev_veh}), the picture inverts. M-TCN collapses under 
tire-temperature corruption, degrading far more sharply than the other models and ending well above them at high severity. This is consistent with M-TCN's worst-on-dataset rPC and its weakest mPC ranking on 
Vehicle Dynamics (Tables~\ref{tab:mpc} and~\ref{tab:rpc}).

\paragraph{Effect of adversarial training.}
On HR (Figure~\ref{fig:sev_ppg_rob}), TST-PGD starts from a 
noticeably worse clean RMSE than vanilla TST but degrades more gradually 
with severity. The clean-data penalty raises TST-PGD's mPC above TST's on 
this dataset; yet rPC, places TST-PGD second on HR (Table~\ref{tab:rpc}), correctly identifying the 
underlying relative robustness gain that mPC obscures. On Vehicle Dynamics 
(Figure~\ref{fig:sev_veh_rob}), nominal performance is comparable between 
the two models, while relative degradation is markedly smaller for TST-PGD 
across most failure modes. With no clean-data penalty to offset, this 
robustness gain registers on both mPC and rPC. The two PGD examples 
together show how rPC and mPC can agree or diverge depending on whether 
robustification incurs a nominal cost.

\section{Statistical significance testing}
\label{app:significance}

We follow the Demšar~\cite{demsar_statistical_2006} protocol as implemented by \texttt{autorank}~\cite{herbold_autorank_2020}: Shapiro--Wilk normality tests select between Bartlett's (normal) or Levene's (non-normal) homogeneity test, which determines whether parametric (ANOVA~+~Tukey HSD) or non-parametric (Friedman~+~Nemenyi) tests are applied ($\alpha = 0.05$). Results are summarized in Table~\ref{tab:significance_summary}; for metrics with a significant omnibus result, we additionally report critical difference (CD) diagrams that visualize mean ranks and non-significant clusters (connected by horizontal bars).

\begin{table}[h]
\caption{Significance results. $N$: observations; $k$: models.}
\label{tab:significance_summary}
\centering
\begin{tabular}{llcccl}
\toprule
\textbf{Metric} & \textbf{Test} & $N$ & $k$ & $p$ & \textbf{Post-hoc} \\
\midrule
nom                              & Friedman & 9 & 9 & $<0.001$ & Nemenyi \\
mPC                              & Friedman & 9 & 9 & $<0.001$ & Nemenyi \\
mPC$_{\mathrm{hf0}}$             & Friedman & 9 & 9 & $<0.001$ & Nemenyi \\
rPC                              & Friedman & 9 & 9 & $<0.001$ & Nemenyi \\
rPC$_{\mathrm{hf0}}$             & Friedman & 9 & 9 & $<0.001$ & Nemenyi \\
$s_{\mathrm{cross}}$             & Friedman & 9 & 9 & $0.187$  & ---     \\
nPC                              & ANOVA    & 9 & 9 & $0.069$  & ---     \\
\bottomrule
\end{tabular}
\end{table}

\begin{figure}[h]
\centering
\includegraphics[width=0.8\linewidth]{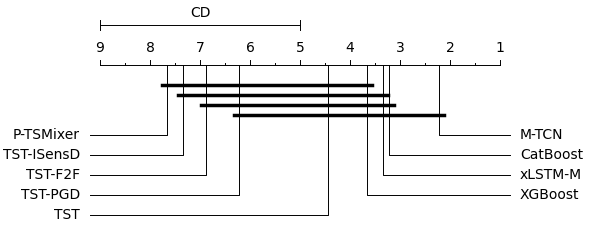}
\caption{Critical difference diagram for nominal performance (nom).
CD$\,{=}\,4.004$.}
\label{fig:cd_nom}
\end{figure}

\paragraph{Nominal Performance}
M-TCN attains the best mean rank, with the tree ensembles, xLSTM-M, plain TST and TST-PGD forming the leading non-significant cluster
(Figure~\ref{fig:cd_nom}); the other robustified TST variants and
P-TSMixer rank significantly worse.

\begin{figure}[h]
\centering
\includegraphics[width=0.8\linewidth]{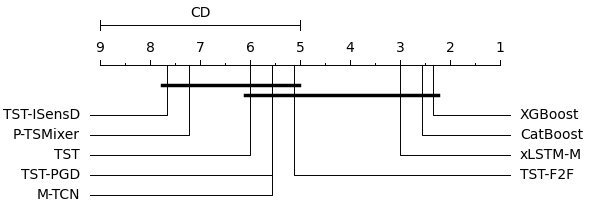}
\caption{Critical difference diagram for mean performance under
corruption (mPC). CD$\,{=}\,4.004$.}
\label{fig:cd_mpc}
\end{figure}

\paragraph{mPC}
XGBoost, CatBoost, and xLSTM-M move to the top, all three significantly outranking P-TSMixer and TST-ISensD
(Figure~\ref{fig:cd_mpc}).  M-TCN, the nominal leader, drops to
mid-pack but remains within the leading non-significant cluster.

\begin{figure}[h]
\centering
\includegraphics[width=0.8\linewidth]{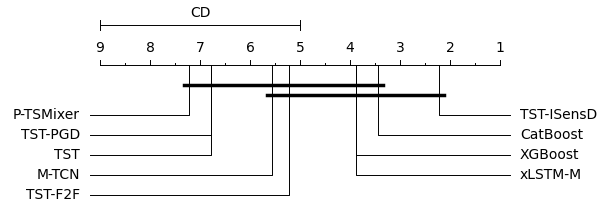}
\caption{Critical difference diagram for mPC restricted to the hard fault with zero severity.
CD$\,{=}\,4.004$.}
\label{fig:cd_mpc_hf0}
\end{figure}

\paragraph{mPC at hard fault with zero severity}
The ranking shifts substantially relative to mPC: TST-ISensD
moves to the top, while plain TST, TST-PGD, and P-TSMixer rank
significantly worse than the leader (Figure~\ref{fig:cd_mpc_hf0}).

\begin{figure}[h]
\centering
\includegraphics[width=0.8\linewidth]{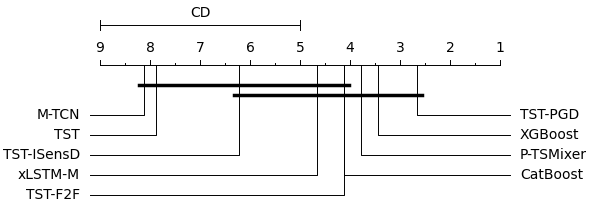}
\caption{Critical difference diagram for relative performance under
corruption (rPC). CD$\,{=}\,4.004$.}
\label{fig:cd_rpc}
\end{figure}

\paragraph{rPC}
TST-PGD, XGBoost, and P-TSMixer lead the ranking; M-TCN ranks last
and plain TST is significantly outranked by all three top models
(Figure~\ref{fig:cd_rpc}). All three robustified TST variants join
the top non-significant cluster, meaning robustification eliminates
TST's significant deficit.

\begin{figure}[h]
\centering
\includegraphics[width=0.8\linewidth]{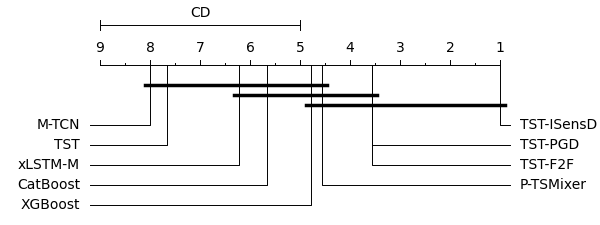}
\caption{Critical difference diagram for rPC restricted to the hard fault with zero severity.
CD$\,{=}\,4.004$.}
\label{fig:cd_rpc_hf0}
\end{figure}

\paragraph{rPC at hard fault with zero severity}
TST-ISensD ranks best by a large margin and is significantly better
than CatBoost, xLSTM-M, plain TST, and M-TCN
(Figure~\ref{fig:cd_rpc_hf0}).

\paragraph{Baseline crossing severity $s_{\mathrm{cross}}$}
Differences are not statistically significant (Friedman, $p=0.187$);
no post-hoc comparisons are performed and no CD diagram is reported.

\paragraph{nPC}
The omnibus test is not significant (ANOVA, $p=0.069$); no post-hoc
comparisons are performed and no CD diagram is reported.
Descriptively, XGBoost, CatBoost, and xLSTM-M rank best, while
P-TSMixer and TST-ISensD rank worst, mirroring the aggregate mPC
ordering. This is expected since the naive baseline rescaling is constant across models for one dataset, therefore
within-dataset ranks are identical to mPC. The discrepancy in
significance arises because the normality-test outcomes route nPC to
ANOVA and mPC to the Friedman test.

\section{Package API reference}
\label{app:api}

This appendix supplements the minimal example in Listing~\ref{lst:muvis-rb-api}
with the details needed to integrate arbitrary models and extend the evaluation
suite.

\paragraph{Data format.}
All input arrays follow an \texttt{(N, T, C)} layout: $N$ samples, $T$ time
steps, $C$ sensor channels. Labels are one-dimensional arrays of shape
\texttt{(N,)}. Data is z-score normalized per channel before any failure mode
is applied.

\paragraph{Registering a model.}
Exactly one of \texttt{predict\_fn} or \texttt{model} must be supplied.
\texttt{predict\_fn} is a callable taking a \texttt{numpy} array of
shape \texttt{(N, T, C)} and returning predictions of shape
\texttt{(N,)}; \texttt{model} is a PyTorch \texttt{nn.Module} or a
scikit-learn estimator with \texttt{.predict}:

\begin{lstlisting}[
  language=Python,
  basicstyle=\small\ttfamily,
  backgroundcolor=\color{gray!5},
  frame=lines,
  framesep=4pt,
  breaklines=true,
  showstringspaces=false,
  columns=fullflexible,
  keepspaces=true,
  tabsize=4,
  commentstyle=\itshape\color{gray!60!black},
  keywordstyle=\bfseries\color{blue!60!black},
  stringstyle=\color{ForestGreen},
]
def predict_fn(X: np.ndarray) -> np.ndarray:
    """X: (N, T, C) -> predictions: (N,)"""
    ...

testbed.add_model("MyFn",  predict_fn=predict_fn)
testbed.add_model("TCN",   model=my_pytorch_module)
testbed.add_model("Ridge", model=sklearn_ridge)
\end{lstlisting}

\paragraph{Custom datasets.}
Datasets can be supplied as a benchmark ID (e.g.\ \texttt{rob.PPGDalia}),
a directory of \texttt{.ts} files, an
\texttt{(X\_train,\ y\_train,\ X\_test,\ y\_test[,\ name])} tuple, or
via \texttt{testbed.add\_dataset(...)}. The helper
\texttt{rob.prepare\_dataset(df,\ target,\ window\_size,\ ...)}
converts a time-ordered \texttt{pandas} DataFrame into the windowed
arrays expected by the testbed.

\paragraph{Custom metrics.}
Subclass \texttt{Metric} and register the instance; \texttt{name}
becomes a column in the raw DataFrame:

\begin{lstlisting}[
  language=Python,
  basicstyle=\small\ttfamily,
  backgroundcolor=\color{gray!5},
  frame=lines,
  framesep=4pt,
  breaklines=true,
  showstringspaces=false,
  columns=fullflexible,
  keepspaces=true,
  tabsize=4,
  commentstyle=\itshape\color{gray!60!black},
  keywordstyle=\bfseries\color{blue!60!black},
  stringstyle=\color{ForestGreen},
]
from muvis_c import Metric

class MAEMetric(Metric):
    name = "mae"
    def compute(self, y_true, y_pred):   # both shape (N,)
        return float(np.mean(np.abs(y_true - y_pred)))

testbed.add_metric(MAEMetric())
\end{lstlisting}

\paragraph{Custom failure modes.}
Subclass \texttt{SeverityFailure} and implement \texttt{apply}, which
returns a corrupted copy of \texttt{X} for severity $s \in [0, 1]$
applied to channel \texttt{feature\_idx}. The base class stores the
maximum perturbation scale \texttt{self.k}:

\begin{lstlisting}[
  language=Python,
  basicstyle=\small\ttfamily,
  backgroundcolor=\color{gray!5},
  frame=lines,
  framesep=4pt,
  breaklines=true,
  showstringspaces=false,
  columns=fullflexible,
  keepspaces=true,
  tabsize=4,
  commentstyle=\itshape\color{gray!60!black},
  keywordstyle=\bfseries\color{blue!60!black},
  stringstyle=\color{ForestGreen},
]
import torch
from muvis_c import SeverityFailure

class SpikeFailure(SeverityFailure):
    """Single spike at the sequence midpoint."""
    def apply(self, X: torch.Tensor, feature_idx: int,
              severity: float) -> torch.Tensor:
        out = X.clone()
        mid = out.shape[1] // 2
        out[:, mid, feature_idx] += severity * self.k
        return out

testbed.add_failure(SpikeFailure(k=5.0))
\end{lstlisting}

\paragraph{Testbed configuration.}
Table~\ref{tab:testbed-params} lists the constructor arguments.

\begin{table}[h]
\centering
\caption{\texttt{Testbed} constructor parameters.}
\label{tab:testbed-params}
\small
\begin{tabular}{@{}llp{6.4cm}@{}}
\toprule
\textbf{Parameter} & \textbf{Default} & \textbf{Description} \\
\midrule
\texttt{dataset}           & \texttt{None}            & Benchmark ID, path to a \texttt{.ts} directory, or \texttt{(X\_train, y\_train, X\_test, y\_test[, name])} tuple. If \texttt{None}, call \texttt{add\_dataset(...)} before \texttt{run()}. \\
\texttt{data\_root}        & \texttt{None}            & Root directory for benchmark IDs; falls back to \texttt{\$MUVIS\_C\_DATA\_DIR} or \texttt{\textasciitilde/.muvis\_c/data/}. \\
\texttt{failures}          & all 10 built-in          & Explicit list of \texttt{SeverityFailure} instances (replaces defaults). Mutually exclusive with \texttt{exclude\_failures}. \\
\texttt{exclude\_failures} & \texttt{None}            & Failure-mode classes to remove from the default set, e.g.\ \texttt{[rob.Outliers]}. \\
\texttt{target\_features}  & \texttt{"all"}           & Feature indices to corrupt. \\
\texttt{severity\_steps}   & \texttt{20}              & Number of severity increments in $[0,1]$ (yields 21 evaluation points). \\
\texttt{k}                 & \texttt{3.0}             & Default maximum perturbation scale passed to built-in failures. \\
\texttt{seed}              & \texttt{42}              & Random seed for reproducibility. \\
\texttt{device}            & \texttt{"auto"}          & \texttt{"cuda"}, \texttt{"mps"}, \texttt{"cpu"}, or auto-detect. \\
\texttt{scaler}            & \texttt{StandardScaler()}& Per-channel scaler fit on the training set; \texttt{None} skips scaling. \\
\texttt{batch\_size}       & \texttt{256}             & Batch size used during the sweep. \\
\bottomrule
\end{tabular}
\end{table}

\paragraph{Results object.}
\texttt{testbed.run()} returns a \texttt{Results} instance wrapping a
single \texttt{pandas} DataFrame (\texttt{Results.raw}) with one row
per (model, dataset, failure mode, feature, severity) tuple. The
DataFrame contains the bootstrap RMSE under corruption
(\texttt{boot\_mean}), the clean reference (\texttt{clean\_boot\_mean}),
the naive mean-predictor reference (\texttt{baseline\_boot\_mean}),
bootstrap confidence intervals, and one column per registered custom
metric. Aggregate quantities are exposed as methods that return
\texttt{model $\times$ dataset} pivot tables
(Table~\ref{tab:results-methods}).

\begin{table}[h]
\centering
\caption{Aggregate metrics on \texttt{Results}. RMSE-derived quantities are
computed on rows with $s>0$ unless noted.}
\label{tab:results-methods}
\small
\begin{tabular}{@{}lp{4.2cm}p{4.6cm}@{}}
\toprule
\textbf{Method} & \textbf{Definition} & \textbf{Reading} \\
\midrule
\texttt{nominal\_rmse()}            & Clean RMSE                                       & Lower is better. \\
\texttt{normalized\_performance()}  & $\mathrm{RMSE}_{\text{clean}}/\mathrm{RMSE}_{\text{baseline}}$ & $<\!1$ beats the naive baseline. \\
\texttt{mpc()}                      & Mean RMSE under corruption (mPC)                 & Same units as RMSE; lower is better. \\
\texttt{npc()}                      & $\text{mPC}/\mathrm{RMSE}_{\text{baseline}}$     & Scale-free; lower is better. \\
\texttt{rpc()}                      & $\text{mPC}/\mathrm{RMSE}_{\text{clean}}$        & $1.0 =$ no degradation. \\
\texttt{rpc\_per\_failure()}        & rPC per failure mode (averaged over datasets)    & Index = failure mode. \\
\texttt{crossing\_severity()}       & Smallest $s>0$ at which any (failure, feature) reaches $\mathrm{RMSE}_{\text{baseline}}$ & Higher is better; \texttt{NaN} = no crossing. \\
\texttt{mpc\_hard\_fault\_zero()}   & mPC restricted to hard-fault rows at $s=0$       & Loss of a single channel. \\
\texttt{rpc\_hard\_fault\_zero()}   & $\text{mPC}_{\text{hf0}}/\mathrm{RMSE}_{\text{clean}}$ & Scale-free counterpart. \\
\bottomrule
\end{tabular}
\end{table}

\noindent For inspection and export, \texttt{Results} additionally
exposes \texttt{summary()} (formatted score table),
\texttt{to\_csv(output\_dir)} (writes
\texttt{output\_dir/per\_severity.csv}), and
\texttt{plot\_degradation(...)} (RMSE-vs-severity curves with optional
filters on dataset, model, failure, or feature).

\paragraph{Multi-dataset evaluation.}
Per-dataset \texttt{Results} objects are combined with the static
\texttt{Results.merge}; the combined object exposes all aggregate
methods above:
\newpage
\begin{lstlisting}[
  language=Python,
  basicstyle=\small\ttfamily,
  backgroundcolor=\color{gray!5},
  frame=lines,
  framesep=4pt,
  breaklines=true,
  showstringspaces=false,
  columns=fullflexible,
  keepspaces=true,
  tabsize=4,
  commentstyle=\itshape\color{gray!60!black},
  keywordstyle=\bfseries\color{blue!60!black},
  stringstyle=\color{ForestGreen},
]
all_results = []
for ds_id in rob.BENCHMARK_DATASETS:
    tb = rob.Testbed(dataset=ds_id)
    tb.add_model("MyModel", predict_fn=load_model(ds_id))
    all_results.append(tb.run())

combined = rob.Results.merge(all_results)
combined.summary()
\end{lstlisting}

\end{document}